\documentclass[letterpaper, 10 pt, conference]{ieeeconf}  

\IEEEoverridecommandlockouts                              

\usepackage{mathptmx} 
\usepackage{amsmath} 
\usepackage{amssymb}  
\usepackage{tikz}
\usepackage{stfloats} 
\usepackage{subcaption}
\usepackage{booktabs}  
\usepackage{hyperref}
\usepackage{url}

\usetikzlibrary{positioning, arrows.meta, shapes.geometric, calc}

\title{\LARGE \bf
Hierarchical Attention and Graph Neural Networks: Toward Drift-Free Pose Estimation}

\author{Kathia Melbouci$^{1}$ and 
Fawzi Nashashibi $^{1}$
\thanks{*This work was carried out in the SAMBA collaborative project, co-funded by BpiFrance in the framework of the Investissement d’Avenir Program (PIA)}
\thanks{$^{1}$ The authors are with The National Institute for Research in Digital Science and Technology (Inria), 2 Rue Simone IFF, 75012 Paris, FRANCE, {\tt\small kathia.melbouci@inria.fr}, {\tt\small fawzi.nashashibi@inria.fr}}
}

\begin{document}

\maketitle
\thispagestyle{empty}
\pagestyle{empty}

\begin{abstract}
The most commonly used method for addressing 3D geometric registration is the iterative closet-point algorithm, this approach is incremental and prone to drift over multiple consecutive frames. The Common strategy to address the drift is the pose graph optimization subsequent to frame-to-frame registration, incorporating a loop closure process that identifies previously visited places.
In this paper, we explore a framework that replaces traditional geometric registration and pose graph optimization with a learned model utilizing hierarchical attention mechanisms and graph neural networks. We propose a strategy to condense the data flow, preserving essential information required for the precise estimation of rigid poses.
Our results, derived from tests on the KITTI Odometry dataset, demonstrate a significant improvement in pose estimation accuracy. This improvement is especially notable in determining rotational components when compared with results obtained through conventional multi-way registration via pose graph optimization. The code will be made available upon completion of the review process.
\end{abstract}
\section{Introduction}
Geometric registration is a critical prerequisite for a myriad of subsequent robotics tasks. These encompass surface reconstruction, obstacle avoidance, augmented reality, sensor fusion and more. The Iterative Closest Point (ICP) algorithm, a key tool for estimating geometric registration for point-clouds, operates by aligning consecutive frames which presuppose an alignment for initialization.

When tracking consecutive frames, errors can gradually build up, resulting in pose drift. This drift is corrected using optimization techniques \cite{melbouci2020lpg}. One such technique includes detecting loop closures, where previously visited places are recognized. The goal is to find the rigid (or similarity) transformation that best re-align the frames \cite{choi2015robust}. Nevertheless, place recognition can lead to false positives, constraining the user to process the optimization part several times to achieve accurate results. Moreover, processing frames with a large number of point clouds to find the optimal matches can be time-consuming, compromising the ability to process in real-time.

In the past few years, several works have investigated the use of new neural network based pipelines to process large point-clouds, essentially attention mechanisms popularized in transformer architectures \cite{vaswani2017attention,https://doi.org/10.48550/arxiv.2207.09238, guo2022attention, khan2022transformers}. 
More specifically, when applied to point-clouds, the attention vector encodes 
the relationship between points, disentangles most relevant information from 
input data \cite{ryoo2021tokenlearner} 
and learn to focus on different aspects of temporal patterns \cite{guo2022attention}. 
This vector is permutation invariant which makes it better candidate for handling unordered point clouds.


In this work, we undertake an examination of the potential for employing a learned model as an alternative to the conventional geometric registration with pose graph optimization. This idea originates from the questions: Can we replace this complex, computationally-intensive process with a unique learned model?.

Our proposed model operates directly on a bundle of frames instead of processing them sequentially. A key feature of this model is a dual-layer hierarchical attention mechanism applied to the concatenated frames. This mechanism enables the model to selectively concentrate on segments of the sequence that are most pertinent to geometric registration. Within the context of our work, this could mean giving more attention to frames that display substantial motion or notable changes in the environment. To distill and capture only the most essential information, we have utilized the maximum values across various dimensions of the point cloud embeddings.

We have evaluated our approach on the KITTI odometry benchmark \cite{Geiger2012CVPR} to demonstrate that this model can leverage large point clouds generated by recent LiDAR sensors, establishing its potential utility in autonomous vehicle tasks.
%

The remainder of this paper is organized as follows: Section \ref{sec:related_work} discusses the work related to attention mechanisms in point cloud registration. In Section \ref{sec:architecture}, we introduce the proposed architecture. Section \ref{sec:experiments} provides details on the experiments we conducted. Finally, Section \ref{sec:Discussion} concludes the paper and outlines the directions for future work.

\section{Related works}
\label{sec:related_work}
For many years, the task of aligning and matching 3D shapes was achieved using classic computer vision techniques. These techniques leverage ICP \cite{924423}, Feature matching using descriptors \cite{5152473}, and RANSAC \cite{zhou2016fast} to remove outliers.

In the past few years, many research works have focused on the development of new neural network based pipelines 
for point clouds registration, following the ground-breaking results obtained with these architectures 
in other domains  
\cite{kendall2015posenet, guo2014deep}.  
However, they faced several challenges: (1) point-sets are irregular and order invariant, (2) scaling up point neural networks to large datasets (N $>$ 10k points) 
remains difficult even with the availability of modern hardwares, 
and (3) the impact of the choice of a specific method 
on the performance of deep learning models 
remains only partially understood \cite{huang2021comprehensive}.

Solutions based on intermediate 
representations such as meshes and voxels were dominant
 until PointNet \cite{qi2017pointnet} which proposed to leverages multiple MLPs and symmetric functions to process unordered point-clouds directly through 3D coordinates, learning both global and local features, to efficiently perform classification and segmentation tasks.

Since then, the attention mechanism popularized in transformer architecture \cite{vaswani2017attention} 
has proven to be a better candidate for handling unordered point clouds.
Indeed, the attention vector is permutation invariant, 
allows encoding long range dependencies and 
thus, it is analogous to a weighted adjacency vector.

It has been demonstrated that predictive performances of attention mechanisms for perception and computer vision tasks can be attributed to their similarities with spacial smoothing: 
(1) they flatten the cost landscapes, (2) they are low-pass filters then less vulnerable to high-frequency noises \cite{park2022vision}. When applied to point clouds, the attention vectors have the ability to encode similarity scores between data points, providing a robust tool to understand complex spatial relationships. Furthermore, using multiple attention heads enables to focus on different aspects of temporal patterns.

As a result, several works have leveraged attention mechanisms to learn 3D geometric registration. The initial step in this process is embedding the point clouds in an appropriate space before introducing them to attention blocks. 
Among these, some works adopt a representation where the point cloud is structured as a graph \cite{wang2019deep, yu2021cofinet, zhang20233d, hu2022nrtnet, slimani2023rocnet}. 
The graph takes the Cartesian coordinates of the data points and their associated normals and/or intensities, and outputs a down-sampled point cloud, with the aggregated nearest-neighbor features. The graph representation enables to process irregular point clouds. However, as the size of the point cloud grows, the computational complexity might increase substantially.

To mitigate time complexity, a variety of approaches learns hierarchical features from the point cloud. PadLoc \cite{arce2023padloc} utilizes the PV-RCNN \cite{shi2020pv} that discretize the point clouds into voxel grids, which are then processed using sparse 3D convolution layers to create a series of feature maps. These maps are combined to generate a bird's eye view (BEV) map from where key points and their associated features are sampled. Lepard \cite{li2022lepard} uses KpConv \cite{thomas2019kpconv} that learns deformable kernel points, enhancing its ability to adapt to the local geometry of the point clouds. 
Super-point based methods \cite{qin2022geometric} divide features from the KpConv into super points where lower layers capture fine-grained local features and higher layers capture more global features.

Other works expand the use of the attention mechanism to non-rigid objects \cite{monji2023review}. The approach detailed in \cite{trappolini2021shape} incorporates raw 3D shapes into the attention block. The attention 
vector has been modified to take into account the points' density in a local area. The method cited in \cite{hu2022nrtnet} use dynamic GCNN architecture \cite{wang2019dynamic} for feature extraction and computes the relative drift of point pairs to address the registration task. To better constraint the registration process, an optimal transport loss is used in \cite{shen2021accurate}.

 
The works cited above predominantly focus on utilizing attention mechanisms on a single frame of data, from which they derive rigid transformation by comparing it to a target frame. In this work, we explore the improvements achieved by applying hierarchical attention mechanisms to bundles of frames simultaneously, aligning with multiway registration techniques. We further explore the combination of attention mechanisms and graph neural networks to understand spatial relationships and dynamics, leveraging both local and global contextual information for pose estimation.

\section{Algorithm architecture}
\label{sec:architecture}
\begin{figure*}[thpb]
\centering
\begin{tikzpicture}[
    node distance=1cm and 0.5cm,
    every node/.style={rectangle, rounded corners, align=center, draw},
    dataset/.style={fill=blue!10},
    preprocess/.style={fill=yellow!10},
    gnn/.style={fill=green!10},
    att/.style={fill=orange!10, opacity=0.7, minimum width=3.15cm, minimum height=1.35cm},
    emptyatt/.style={fill=orange!10, opacity=0.7, text opacity=0, minimum width=3.15cm, minimum height=1.35cm},
    fusion/.style={fill=purple!10},
    decoder/.style={fill=red!10},
    arrow/.style={-Latex, thick}
]

\node[preprocess] (preprocess) {Point clouds \\ Preprocessing};
\node[gnn, right=of preprocess] (gnn) {Graph\\Neural\\Network};

\node[emptyatt, right=0.8cm of gnn, yshift=0.36cm, xshift=0.18cm] (att3) {};
\node[emptyatt, at=(att3), yshift=-0.18cm, xshift=-0.18cm] (att2) {};
\node[att, at=(att2), yshift=-0.18cm, xshift=-0.18cm] (att1) {Multi-Head\\Self Attention};

\node[circle, draw, minimum size=15pt, right=0.8cm of att2, yshift=-0.18cm, xshift=-0.18cm, label=below:{Feature\\Selection}] (fusion) {};

\node[emptyatt, right=0.8cm of fusion, yshift=0.36cm, xshift=0.18cm] (att6) {};
\node[emptyatt, at=(att6), yshift=-0.18cm, xshift=-0.18cm] (att5) {};
\node[att, at=(att5), yshift=-0.18cm, xshift=-0.18cm] (att4) {Multi-Head\\Cross Attention};

\node[decoder, right=of att5, yshift=-0.18cm, xshift=-0.08cm] (decoder) {6DoF Poses\\Decoder};

\draw[arrow] (preprocess.east) -- (gnn.west);
\draw[arrow] (gnn.east) -- (att1.west);
\draw[arrow] (att1.east) -- (fusion.west);
\draw[arrow] (fusion.east) -- (att4.west);
\draw[arrow] (att4.east) -- (decoder.west);
\end{tikzpicture}
\caption{Graphical Overview of the Ego-Pose Estimation Framework: The pose estimation model takes $N$ point-clouds as input, with each point cloud represented through a distinct set of embeddings. These embeddings are learned using a Graph Neural Network (DGCNN), which processes them to obtain an enriched representation for each point cloud. These representations are then passed through a hierarchical attention mechanism, that operates in two distinct stages: self-attention and cross-attention. These mechanisms help to capture of both intra and inter-set relationships respectively. Derived self attention scores guide the salient features' selection, carrying aggregated information to be utilized by the subsequent cross-attention layer, enhancing the comprehensibility of global context in each point cloud. The 6DoF poses decoder leverages this rich representation to deduce the rigid transformations needed to align the point clouds in a common reference frame, guided by the loss function defined in Equation. (\ref{eq:loss-function}).}
\label{fig:modelUpdated}
\end{figure*}
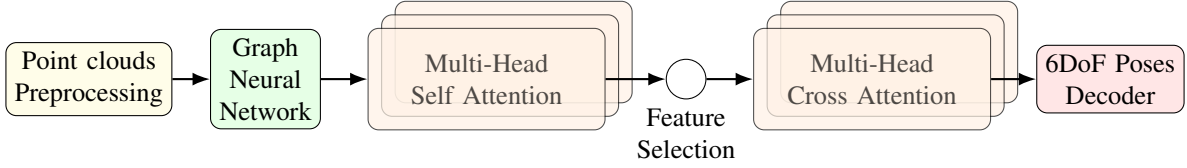

An overview of our model is shown in Figure \ref{fig:modelUpdated}.

Let \( X \) represent the collection of LiDAR scans, denoted as \( X = \{ X_1, X_2, \ldots, X_n \} \), where each point cloud \( X_i = \{p_1, p_2, \ldots, p_M\} \) is of shape \([M, D]\). Here, \( M \) refers to the number of points $p_i$ in each scan, and \( D \) indicates the dimensionality of the points. For 3D cartesian coordinates, \(D=3\), but \(D\) could be higher if more coordinates are added.
We are interested in finding a set of \( n \) rigid transformations, \(\{ T_1, T_2, \ldots, T_n \}\), from the special Euclidean group \(SE(3)\), to align each \( X_i \) to a common reference frame \( F_w \). Each transformation \(T_i\) is expressed as a $(4\times 4)$ matrix including a $(3 \times 3)$ rotation matrix \(R_i\) from the special orthogonal group \(SO(3)\) and a $(3 \times 1)$ translation vector \(t_i\).

Without loss of generality, we consider  \( F_w \)  to be identical to \( X_1 \) for the remainder of the paper.


\paragraph*{\textbf{Point cloud encoder}}The initial phase of our architectural framework focuses on generating embeddings for each data point in a given point cloud, denoted as \(E \equiv \{e_1, e_2, \ldots, e_M\}\). 
\[
e_i = \text{EmbeddingFunction}(p_i), \quad \forall p_i \in X_i
\]
To achieve this we employ a variation of the Dynamic Graph CNN (DGCNN) architecture\cite{wang2019dynamic}. The first step consists of mapping the point normals into a higher-dimensional feature space through the application of dual convolutional layers. These enriched normal features are combined with the original data points to serve as the input for the DGCNN.
This later enables the generation of node representations that integrate both topological connectivity and feature attributes from localized neighborhoods \cite{guo2021pct}.

\paragraph*{\textbf{Hierarchical Attention Mechanism}} In the second step, we introduce a hierarchical attention mechanism. This consists of dual-stage attention, where the first stage -multi-head self attention in Figure \ref{fig:modelUpdated}- focuses on quantifying relationships among points within individual point cloud. It accomplishes this by calculating attention scores based on the embeddings generated for each data point.
\[
A_{ij}^{(1)} = \text{softmax}(f(e_i, e_j)), \quad \forall e_i, e_j \in E
\]
\[
e_i' = \sum_{j} A_{ij}^{(1)} \cdot e_j
\]
where \(f(\cdot, \cdot)\) is a function that computes a normalized attention score between the features embeddings \(e_i\) and \(e_j'\).

To extract the most salient features from a \(D'\)-dimensional space, spanning \(M'\) points within each point cloud (where \(D'\) and \(M'\) are derived from the point cloud encoder), we use a maximum operation to obtain the top \(S\) embeddings. 

\[
e''_i = \left\{ e'_{i\sigma_1}, e'_{i\sigma_2}, \ldots, e'_{i \sigma_S} \right\}
\]
\[
\forall i \in \{1, 2, \ldots, M'\}
\]

Where \(S\) is the predefined number such that \(1 \leq S \leq D'\), and \(\sigma\) is a permutation function that sorts the indices of vector \(e_i'\) in decreasing order based on their corresponding values in vector \(e_i'\).

%


The second stage of attention -multi-head cross attention in Figure \ref{fig:modelUpdated}- is designed to encode relationships between points across different sets. This enables our model providing a comprehensive representation that accounts for global context within the point cloud data.

\[
A_{kl}^{(2)} = \text{softmax}\left(g(e_k'', e_l'')\right) \quad \text{for} \quad k \neq l
\]
where  \(g(\cdot, \cdot)\) is a function that computes a normalized attention score between the features \(e_k''\) and \(e_l''\).

\[
e_k''' = \sum_{l \neq k} A_{kl}^{(2)} \cdot e_l''
\]

\paragraph*{\textbf{Rigid Transformation Decoder}}
To align each point cloud with the global coordinate system, we utilize a Multilayer Perceptron (MLP) on the \(N \times M'\) embeddings derived from the cross-attention module. The initial frame is set as a $(4 \times 4)$ identity matrix. The MLP is designed to generate unique feature sets representing both translation and rotation. For translation, it outputs three features, each corresponding to the x, y, and z coordinates.

The rotation representation, however, depends on the specific parametrization approach selected. In this work, we adopt the Gram-Schmidt orthonormalization method \cite{zhou2019continuity}. This method is used to project the rotation features outputted by the decoder onto the nearest rotation matrix. In utilizing the Gram-Schmidt orthonormalization, the output rotation of the MLP encompasses a 6-dimensional space, reshaped as $(3 \times 2)$ matrix $W$ . The nearest rotation matrix is then retrieved by minimizing the Frobenius norm as illustrated in Eq. \ref{eq:special-gramchmit}.

\begin{equation}
\text{argmin}_{R^* \in SO(3)} \Vert R^* - W_{3\times2} \Vert^2_F,
\label{eq:special-gramchmit}
\end{equation}
%
%
%
%
%

Nonetheless, it is possible to use the special orthogonal procrustes parametrization as presented by Bregier et al. \cite{bregier2021deep}, which includes a total of 9 features for the rotation. The arbitrary $(3 \times 3)$ matrix $W$ produced by the decoder is then projected onto the nearest rotation matrix by minimizing the Frobenius norm (Eq. \ref{eq:special-procrustes}).

\begin{equation}
\text{argmin}_{R^* \in SO(3)} \Vert R^* - W_{3\times3} \Vert^2_F,
\label{eq:special-procrustes}
\end{equation}
%

\paragraph*{\textbf{Loss functions}}
We formulate the loss function with individual constraint on the translation and rotation components. Since estimating distances tends to be simpler than directly determining absolute poses, we define the loss function as depicted in Eq. (\ref{eq:loss-function}).

\begin{equation}
\mathcal{L}(\theta) = \alpha f_t(\Delta t^*_{ij}, \Delta t_{ij}) + \beta f_R(\Delta R^*_{ij}, \Delta R_{ij}),
\label{eq:loss-function}
\end{equation}

Where \( \alpha \), \( \beta \) are weighting factors to balance the contributions of the translational and rotational components respectively. \( f_t \)  returns the mean squared error (MSE) between the pair ($\Delta t^*_{ij}, \Delta t_{ij}$) and \( f_R \) returns the angular distance $\delta$ between a pair of rotation matrices, based on the equality $\mid \Delta R^*_{ij} - \Delta R_{ij} \mid_F = 2\sqrt{2}sin(\delta /2)$.

\section{Experiments}
\label{sec:experiments}
To evaluate the efficacy of our architecture, we benchmarked it against the multiway registration with pose graph optimization as outlined in \cite{zhou2016fast} and available in the Open3D library \cite{Zhou2018}. We evaluated our model using the KITTI odometry dataset \cite{Geiger2012CVPR} to demonstrate its applicability in scenarios involving a large number of points.

In multiway registration via pose graph optimization a pose graph is created where each node represents a piece of geometry associated with a pose matrix transforming it into a global space. These nodes are linked by edges that signify the relationship and transformations needed to align one piece of geometry with another. The alignments are categorized into two classes: odometry edges for neighboring nodes aligned using local registration techniques like ICP, and loop closure edges for non-neighboring nodes - considered less reliable- aligned through global registration.\\
To optimize the pose graph, uncertain parameters and information matrices are assigned to the edges, allowing for the identification and pruning of false alignments through a two-pass optimization process. The first pass tries to identify false alignments considering all edges, while the second pass optimizes the alignment without these false alignments.\\ Different optimization methods and criteria can be selected to suit specific needs and achieve better results. In our experiment we have fixed the voxel size to be 0.05 and we kept the default values -as specified in Open3D- for the remaining parameters \cite{Zhou2018}.

\begin{figure}[h]
  \begin{minipage}[b]{0.24\textwidth}
    \includegraphics[width=\textwidth]{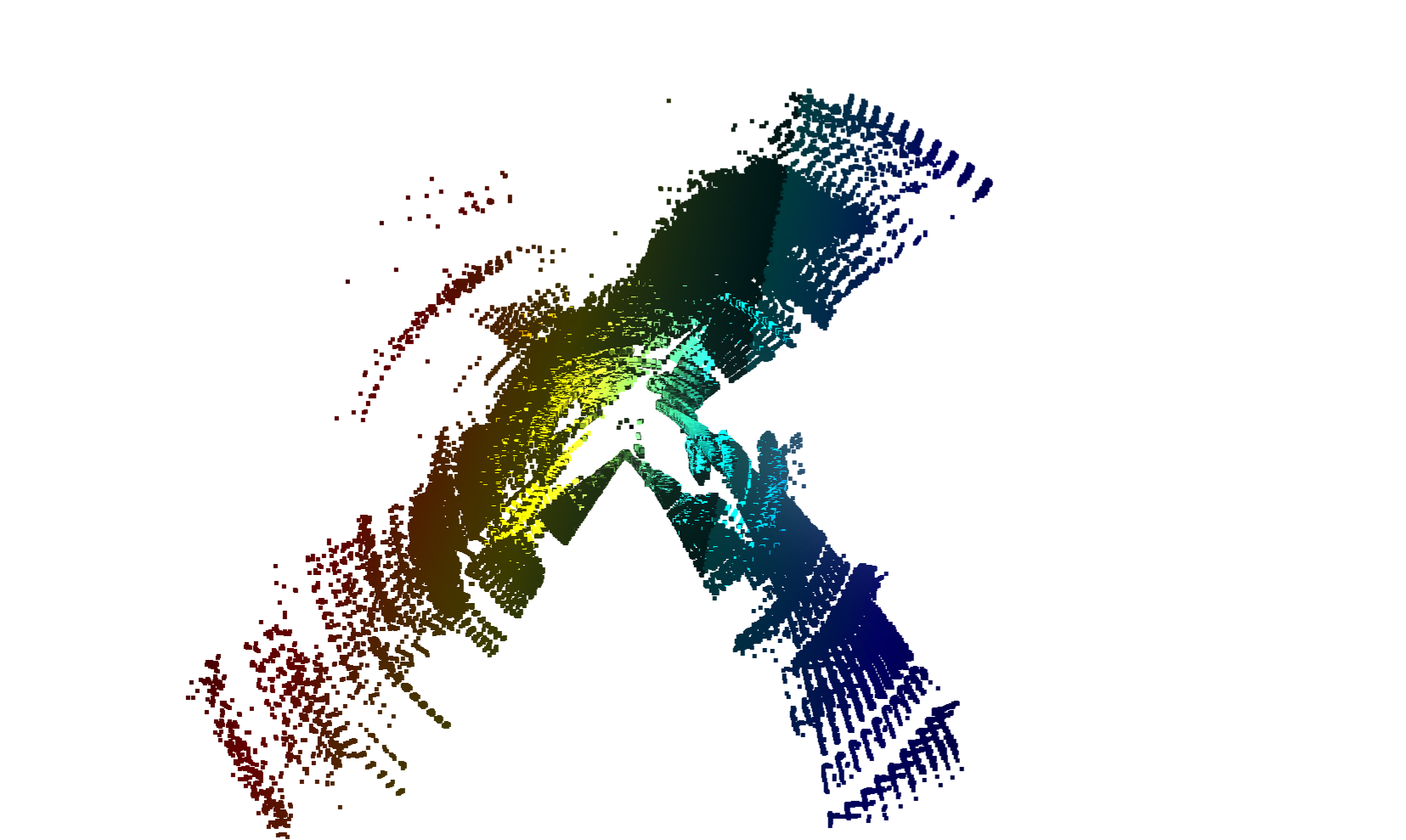}
  \end{minipage}
  \hfill
  \begin{minipage}[b]{0.24\textwidth}
    \includegraphics[width=\textwidth]{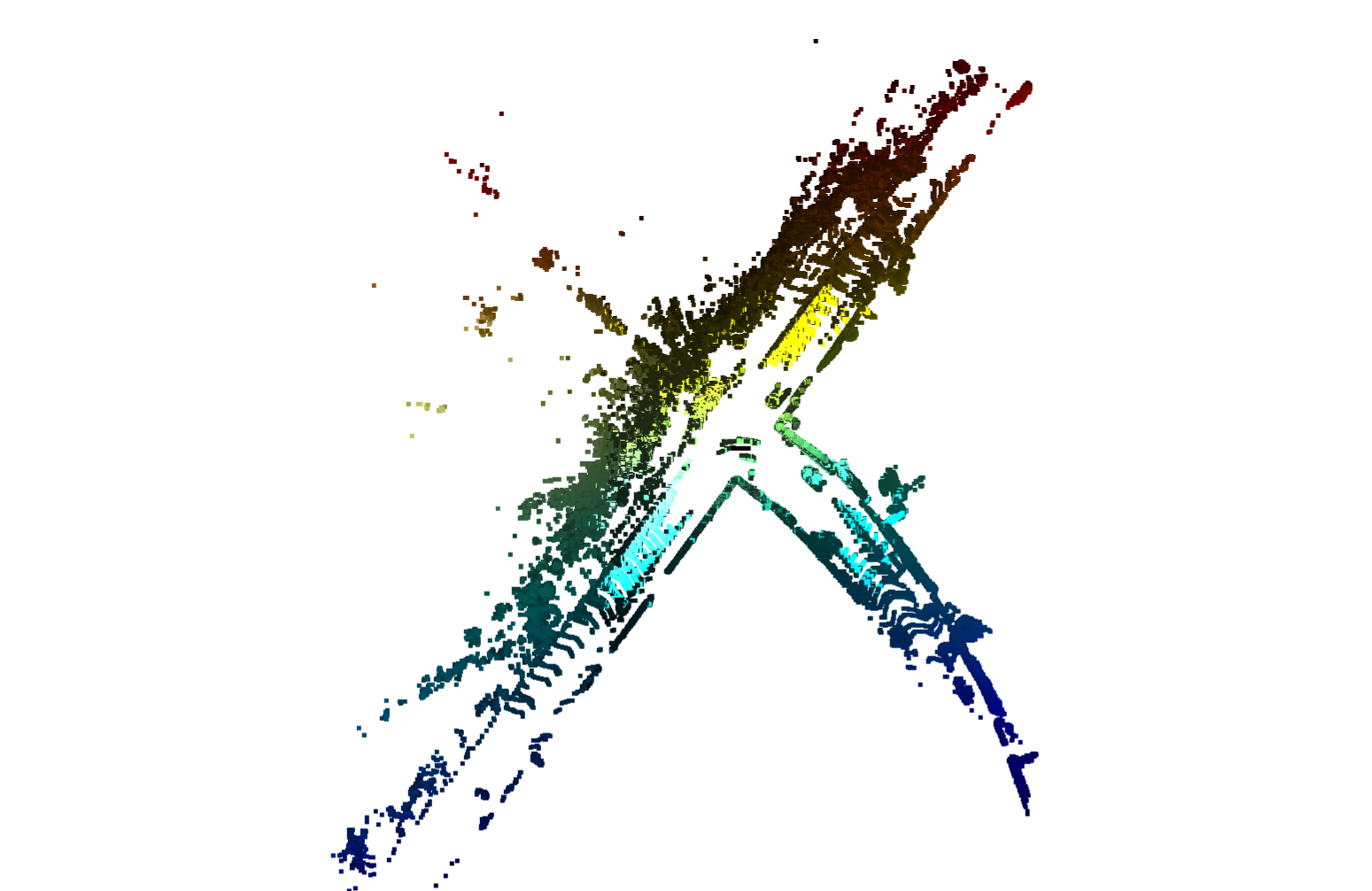}
  \end{minipage}
  \caption{Left: Ten non-aligned scans from Sequence 7 of KITTI odometry benchmark; Right: aligned point clouds using our model-derived poses.}
  \label{fig:aligned-and-non-aligned-pcd}
\end{figure}

The training of our model is done on single NVIDIA GeForce RTX 2080 Ti with $\sim{10G}$ memory. We chose to train our model on a one-second window of a specific trajectory, equivalent to roughly $\sim{10}$ LiDAR scans. These scans are fed sequentially into the model, which outputs a corresponding sequence of rigid poses (Figure \ref{fig:aligned-and-non-aligned-pcd}). It's worth noting that the window parameter, once set during training, can be reduced for inference without requiring retraining. However, increasing the window parameter necessitates specifying the new value and retraining the model from scratch.

\begin{figure}[h]
  \begin{minipage}[b]{0.24\textwidth}
    \includegraphics[width=1.0\textwidth]{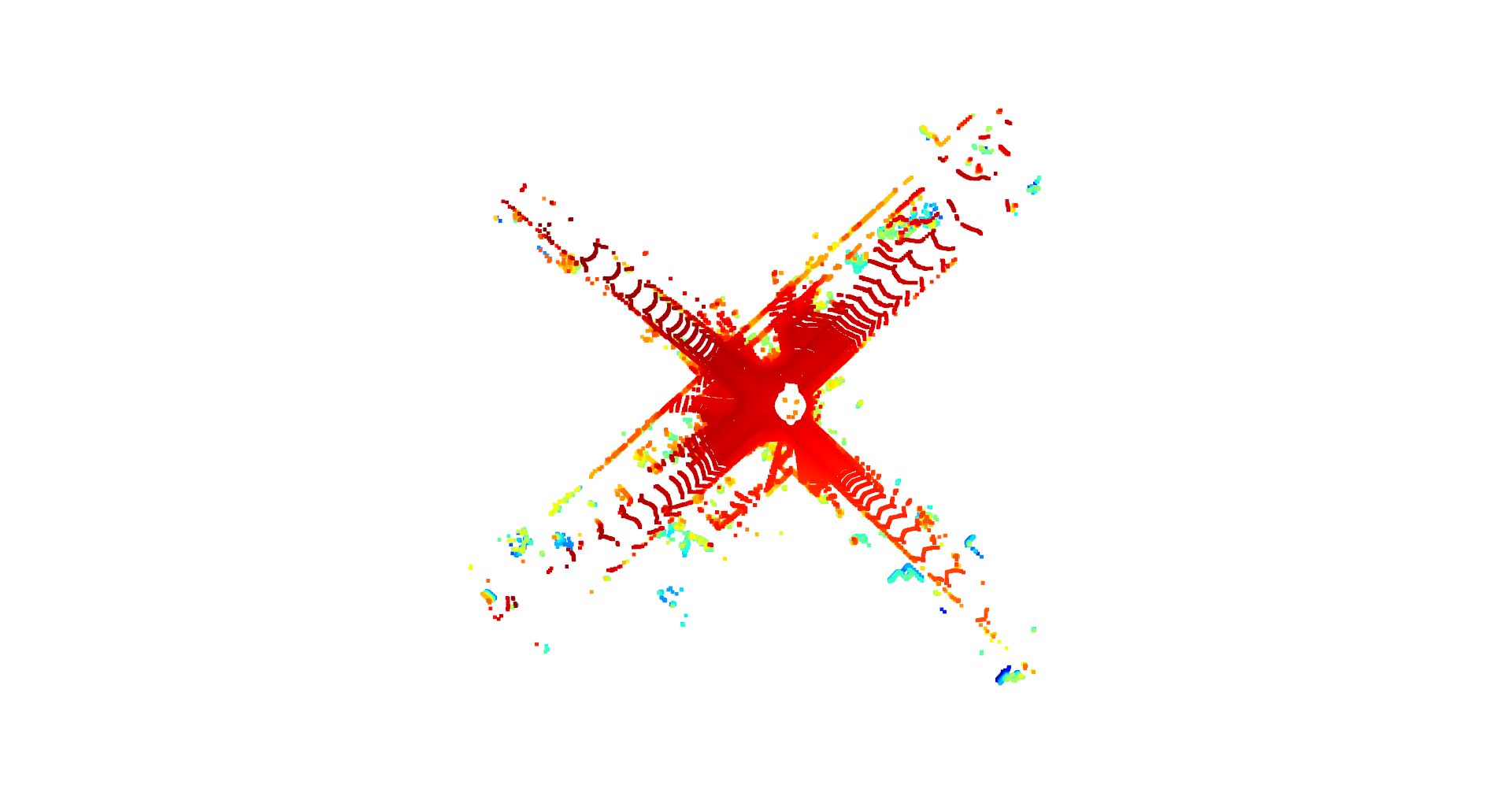}
    \caption*{(a)}
  \end{minipage}
  \hfill
  \begin{minipage}[b]{0.24\textwidth}
    \includegraphics[width=1.0\textwidth]{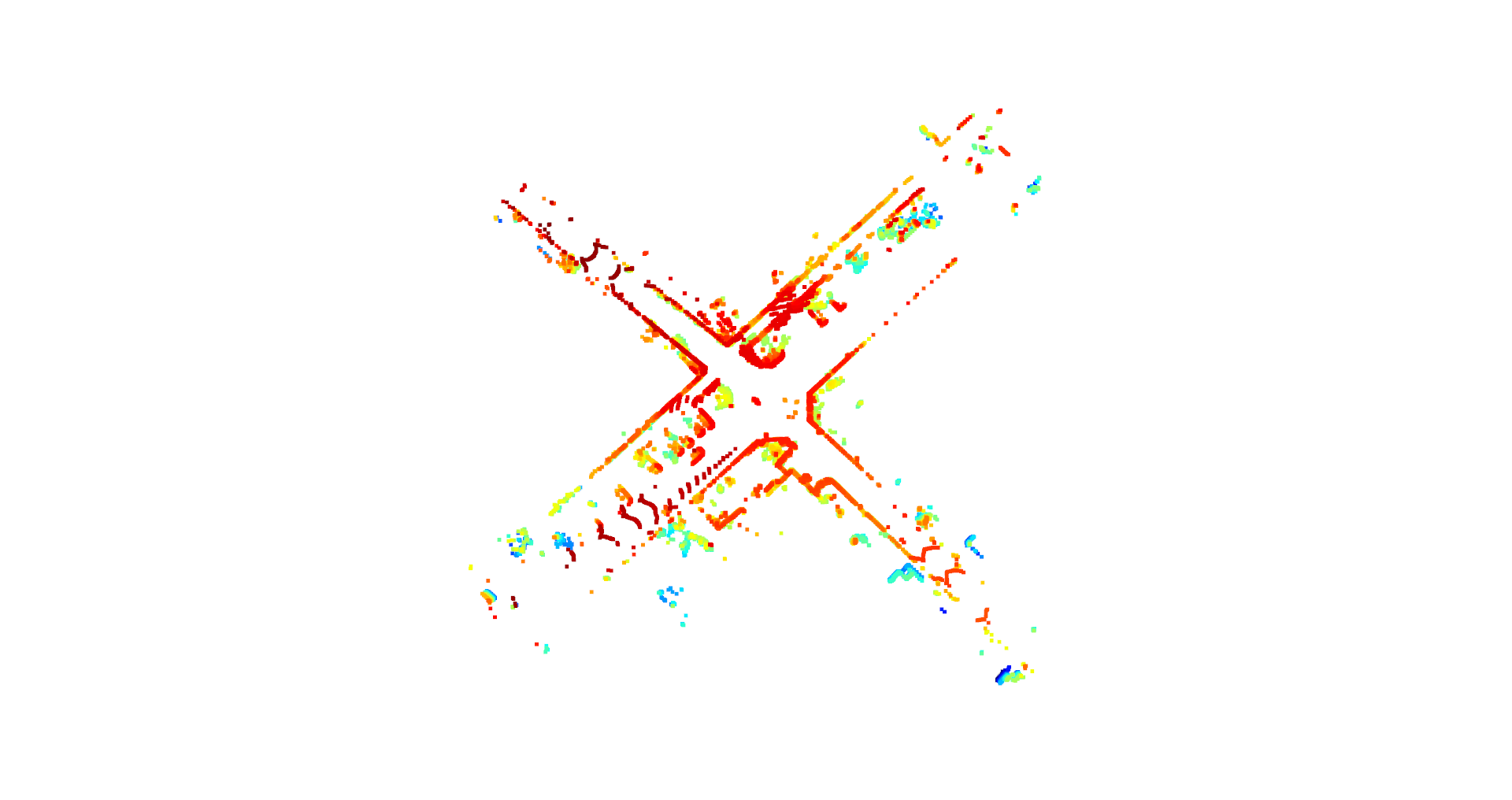}
    \caption*{(b)}
  \end{minipage}
  \caption{Visual comparison of point cloud data from KITTI odometry sequence 5. (a) shows the original, unprocessed point cloud, while (b) displays the point cloud after segmentation to remove rough ground features and reduce the number of points.}
  \label{fig:raw-segmented-pcd}
\end{figure}

\paragraph*{\textbf{Point cloud preprocessing}}
Before feeding the point cloud data into the graph neural network, we subject it to a series of preprocessing steps. First, we employ voxel downsampling to reduce the data's complexity by decreasing the number of points. Next, we employ the RANSAC algorithm to fit a plane that separates ground features, thereby removing any undesired rough ground surfaces from the data. These procedures are illustrated in Figure \ref{fig:raw-segmented-pcd}. \\
Upon completion of these preprocessing steps, each processed point cloud contains approximately $70,000$ points. To standardize the tensor sizes, padding is applied to each point cloud to match the dimensions of the largest point cloud.



\paragraph*{\textbf{Rigid pose estimation}}
Each point cloud in the sequence is transformed into vectors with dimensions \([1024, 256]\) through the DGCNN network \cite{wang2019dynamic}. These vectors serve as inputs to determine attention scores in the first stage of the attention mechanism.

To aggregate this high-dimensional data while retaining the most significant features, we use a max pooling operation. This yields a one-dimensional vector for each point cloud in the batch, capturing the most important attributes from the initial set of 1024 points.

By highlighting the peak value in each feature dimension, we aim to accentuate the most noteworthy characteristics of each point cloud. This leads to a concise yet data-rich 1024-dimension representation for every cloud in the batch, allowing for the preservation of key attributes.

The sequence, configured as [10, 1024], is subsequently processed by an additional attention mechanism to discern the interrelations between distinct point clouds. In our experiments, we've configured both attention mechanisms with 4 heads and 4 layers.

The 6DOF pose decoder, configured as a two-stage Multi-Layer Perceptron (MLP), generates 9-dimensional feature set to represent both translation and rotation. 
The translation component is represented through three outputs corresponding to the x, y, and z coordinates. Meanwhile, the rotation is represented using six features to project the arbitrary $(3 \times 2)$ matrix onto a rotation matrix using special Gram-Schmidt orthonormalization \cite{zhou2019continuity}, as explained in section \ref{sec:architecture}. For this task, we leveraged the Roma library \cite{bregier2021deep},  maintaining a value of 1 for both \(\alpha\) and \(\beta\) parameters.

The qualitative results are presented in Figure \ref{fig:pcd-comparison}. This Figure showcases the alignment of point clouds using three methods: the ground truth, the multiway registration via pose graph optimization, and our proposed approach. These alignments are demonstrated for select examples from the test trajectories from the KITTI Odometry benchmark \cite{Geiger2012CVPR}.

The quantitative results are detailed in Table \ref{tab:are-errors}. This table presents the average Root Mean Square Error (RMSE) values derived from absolute pose errors, showcasing the performance of our model over various sequences. These averages have been computed across 100 sub-trajectories, with each trajectory containing 10 scans. The RMSE values are differentiated into translational (measured in meters) and rotational (measured in radians) errors to offer a comprehensive view of the trajectory accuracies observed in the different sequences

\begin{table*}
    \centering
       \caption{Comparison of RMSE for absolute pose error in translation and rotation.}

    \label{tab:results}
    \begin{tabular}{lcccccc}
        \toprule
        Sequence & \multicolumn{2}{c}{Our Model} & \multicolumn{2}{c}{Multiway registration via Pose Graph Optimization} \\
        & RMSE Trans(m). & RMSE Rot(rad). & RMSE Trans(m). & RMSE Rot(rad). \\
        \midrule
        Sequence 0 &  \textbf{0.064} & \textbf{0.614} & 0.385 & 2.842  \\
        Sequence 1 &  \textbf{0.027} &  \textbf{0.016} & 0.068 & 0.065 \\
        Sequence 2 &  \textbf{0.363} &  \textbf{0.121} & 0.407 & 1.056 \\
        Sequence 3 &  \textbf{0.027} &  \textbf{0.016} & 0.068 & 0.065 \\
        Sequence 4 &  \textbf{0.181} &  \textbf{0.745} & 0.249 &  2.420 \\
        Sequence 5 &  0.051 & \textbf{0.013} & \textbf{0.017} & 0.201 \\
        Sequence 6 &  \textbf{0.387} &  \textbf{1.851} & 2.158 & 2.473 \\
        Sequence 7 &  \textbf{0.013} & \textbf{0.018} & 0.105 & 0.098 \\
        Sequence 8 &  \textbf{0.384} & \textbf{0.689} & 3.234 &  0.507 \\
        Sequence 9 &  \textbf{0.192} & \textbf{0.369} &  0.443 &  0.521 \\
        \bottomrule
    \end{tabular}
    \label{tab:are-errors}
\end{table*}

%
%
\begin{figure*}[t]
  \begin{minipage}[b]{0.24\textwidth}
    \centering
    \includegraphics[width=0.99\linewidth]{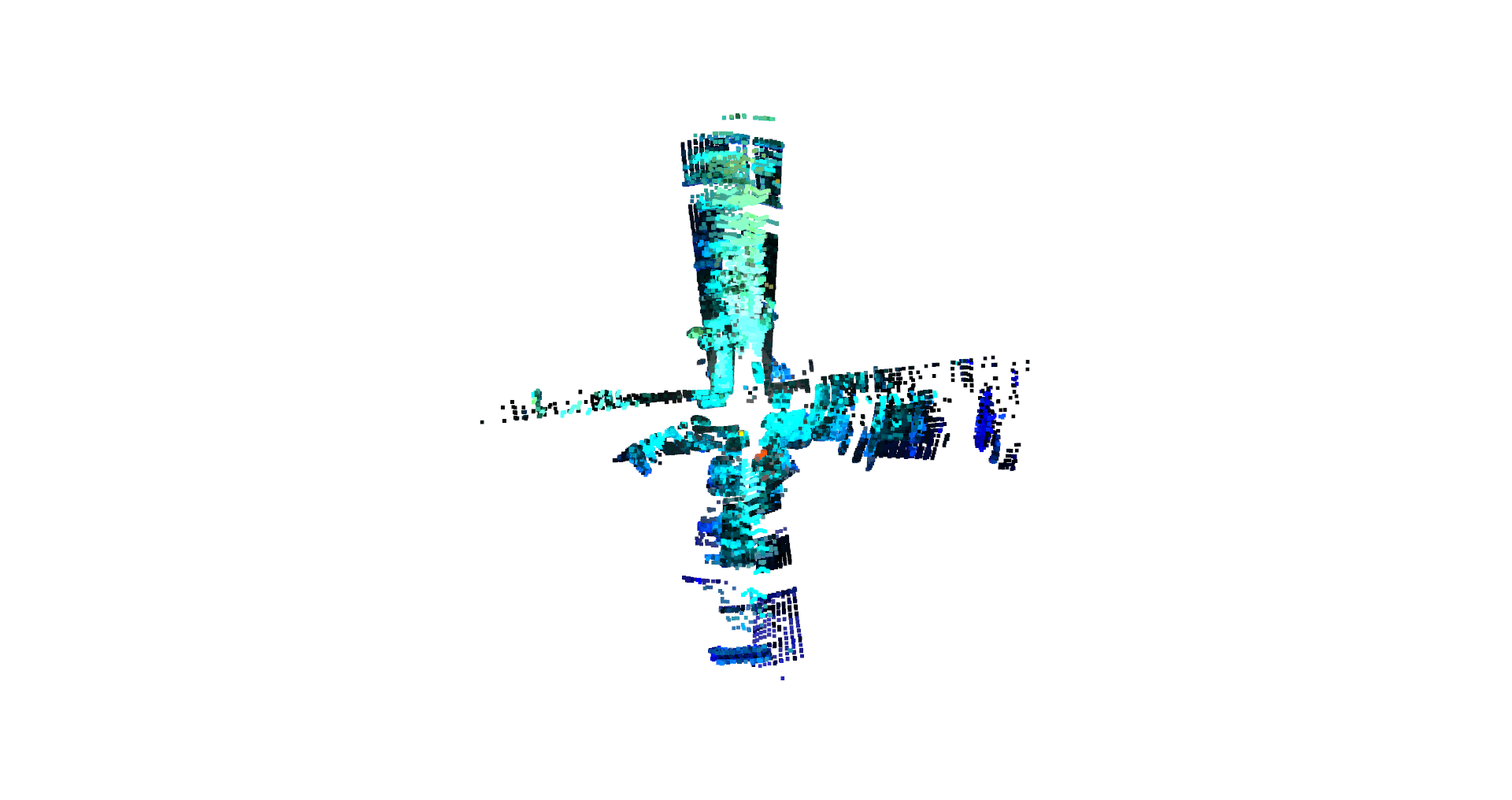}
    \vspace{0.1cm} 
    \includegraphics[width=0.99\linewidth]{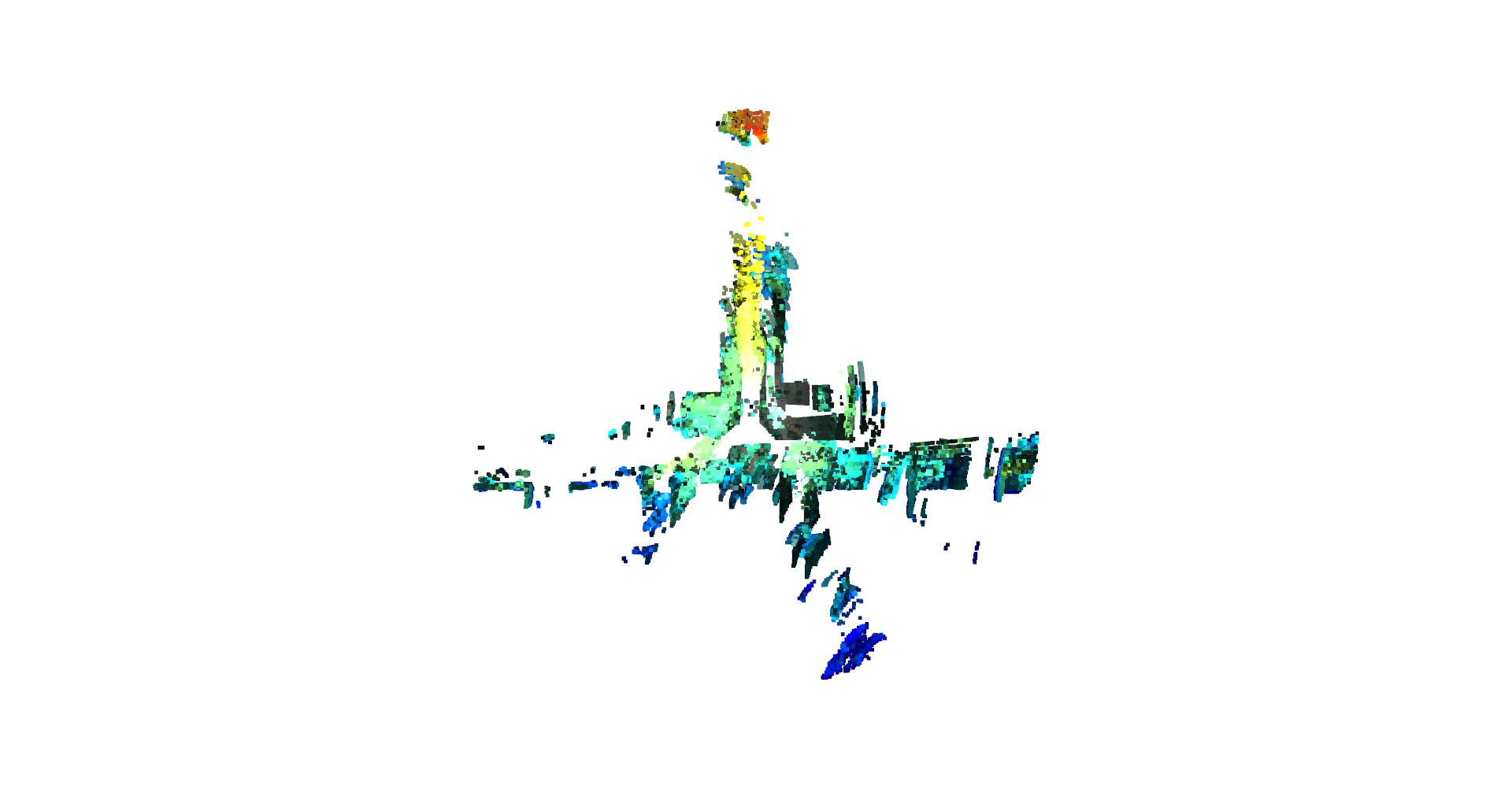}
    \vspace{0.1cm} 
    \includegraphics[width=0.99\linewidth]{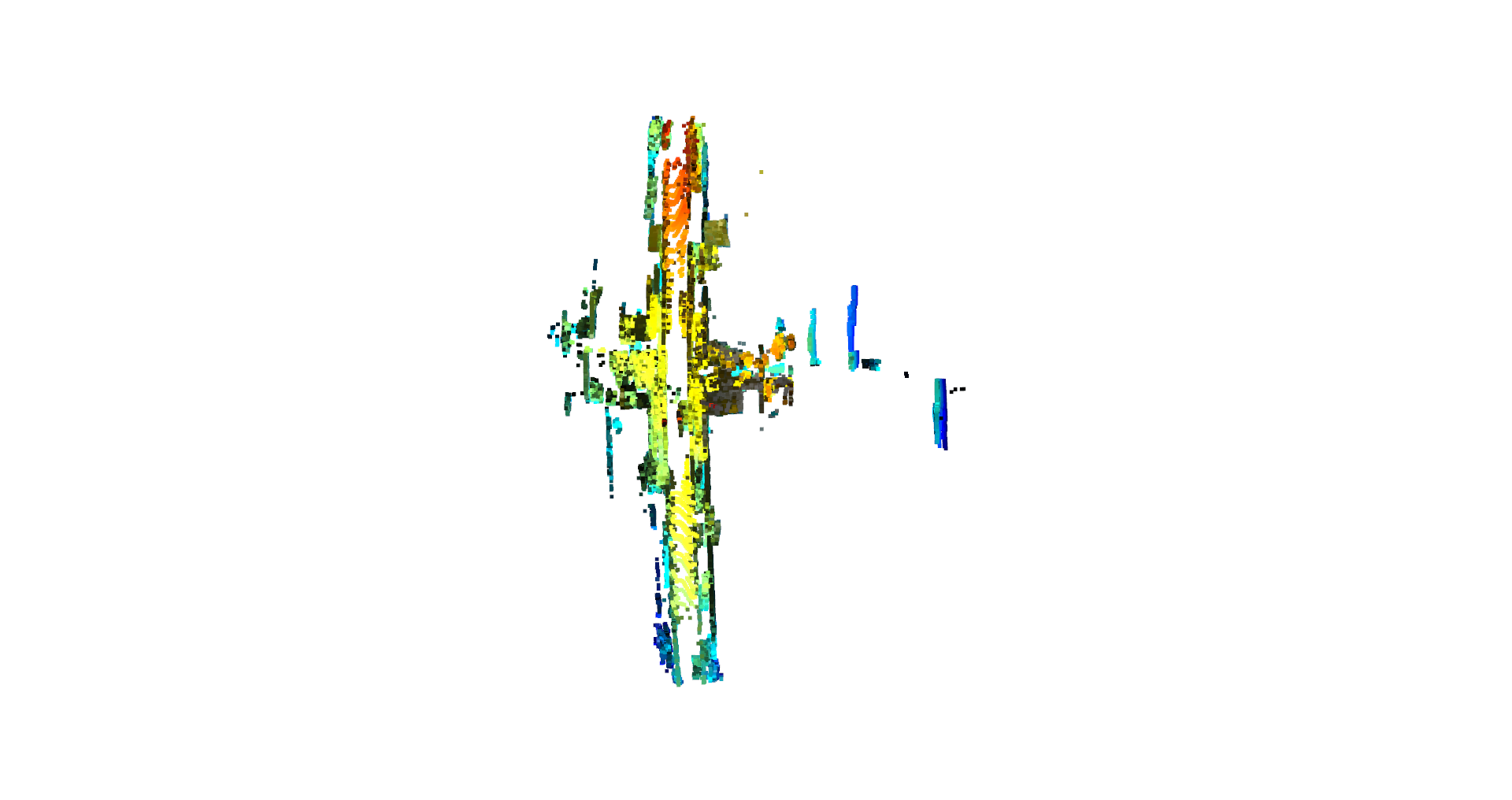}
    \vspace{0.1cm} 
    \includegraphics[width=0.99\linewidth]{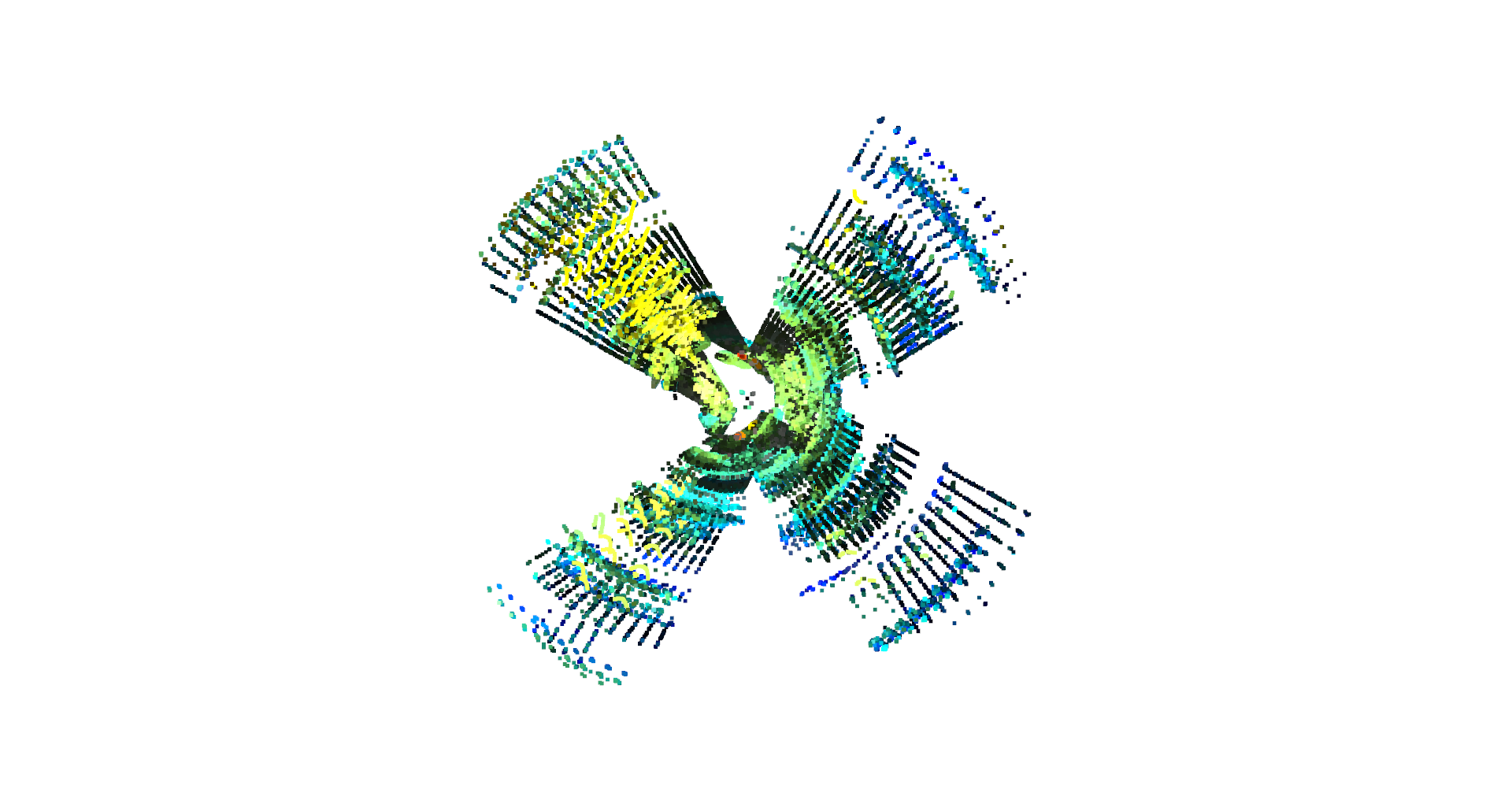}
     \vspace{0.1cm} 
    \includegraphics[width=0.99\linewidth]{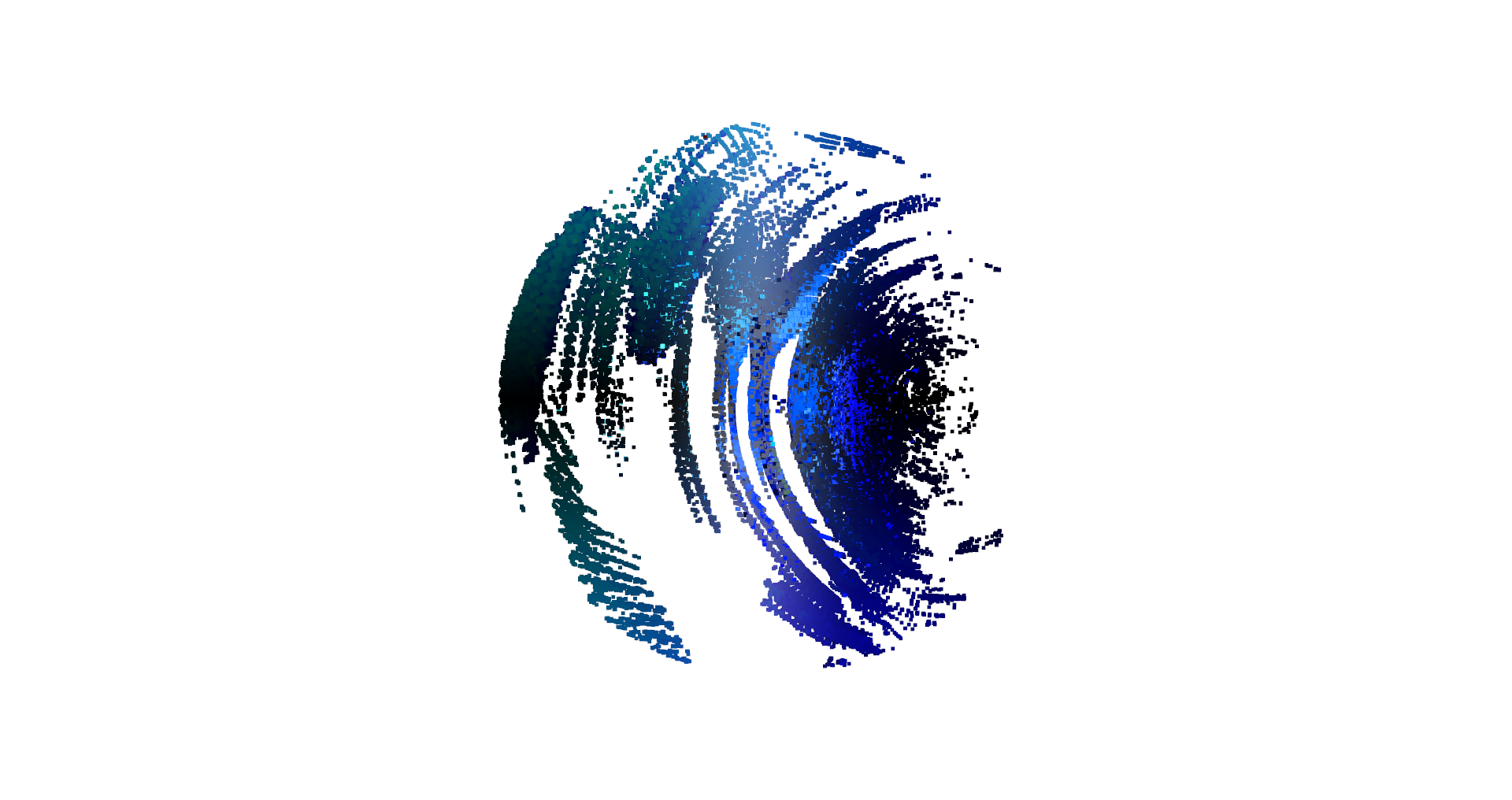}    
  \end{minipage}
  \hfill
  \begin{minipage}[b]{0.24\textwidth}
    \centering
    \includegraphics[width=0.99\linewidth]{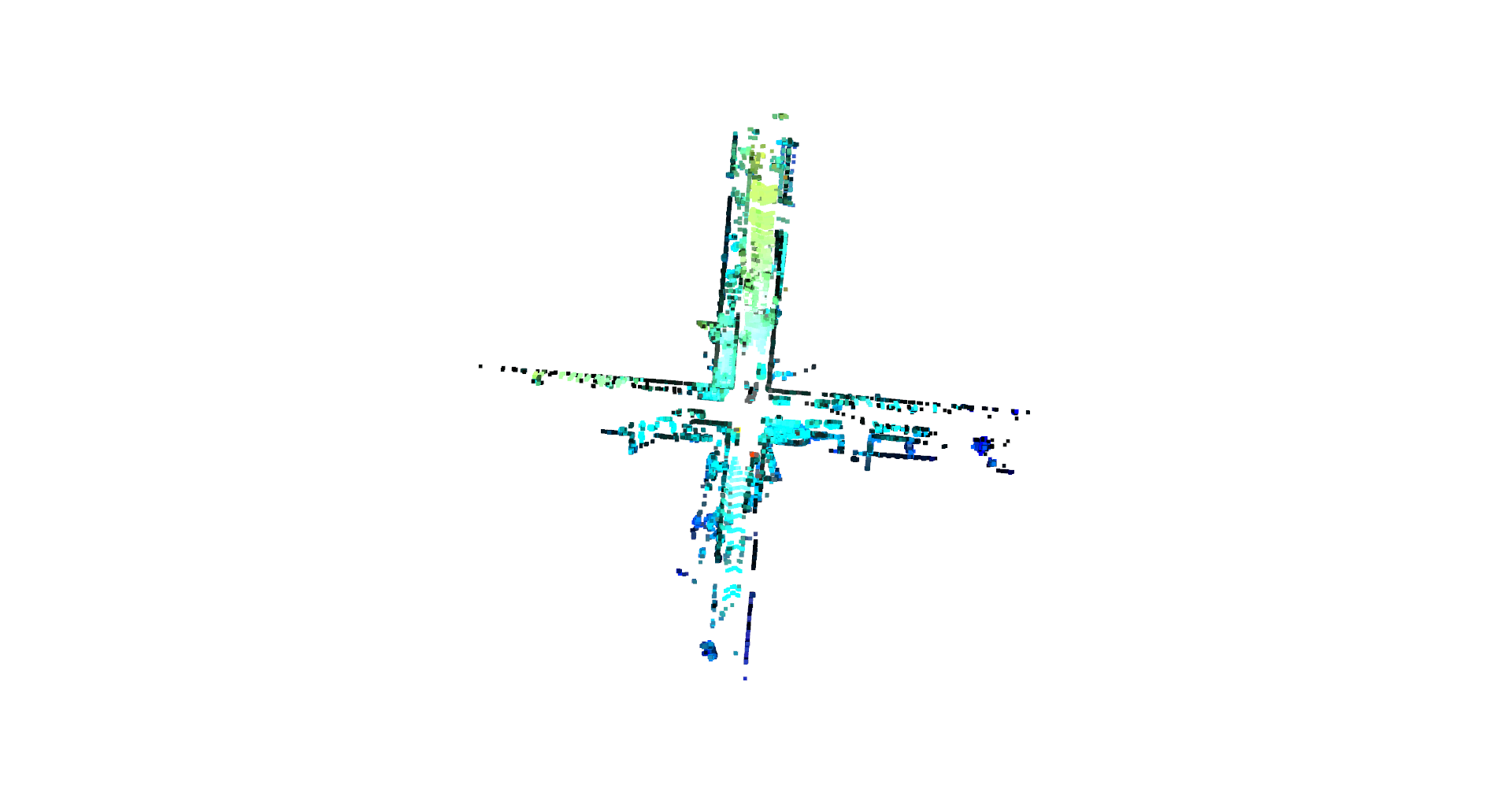}
    \vspace{0.1cm} 
    \includegraphics[width=0.99\linewidth]{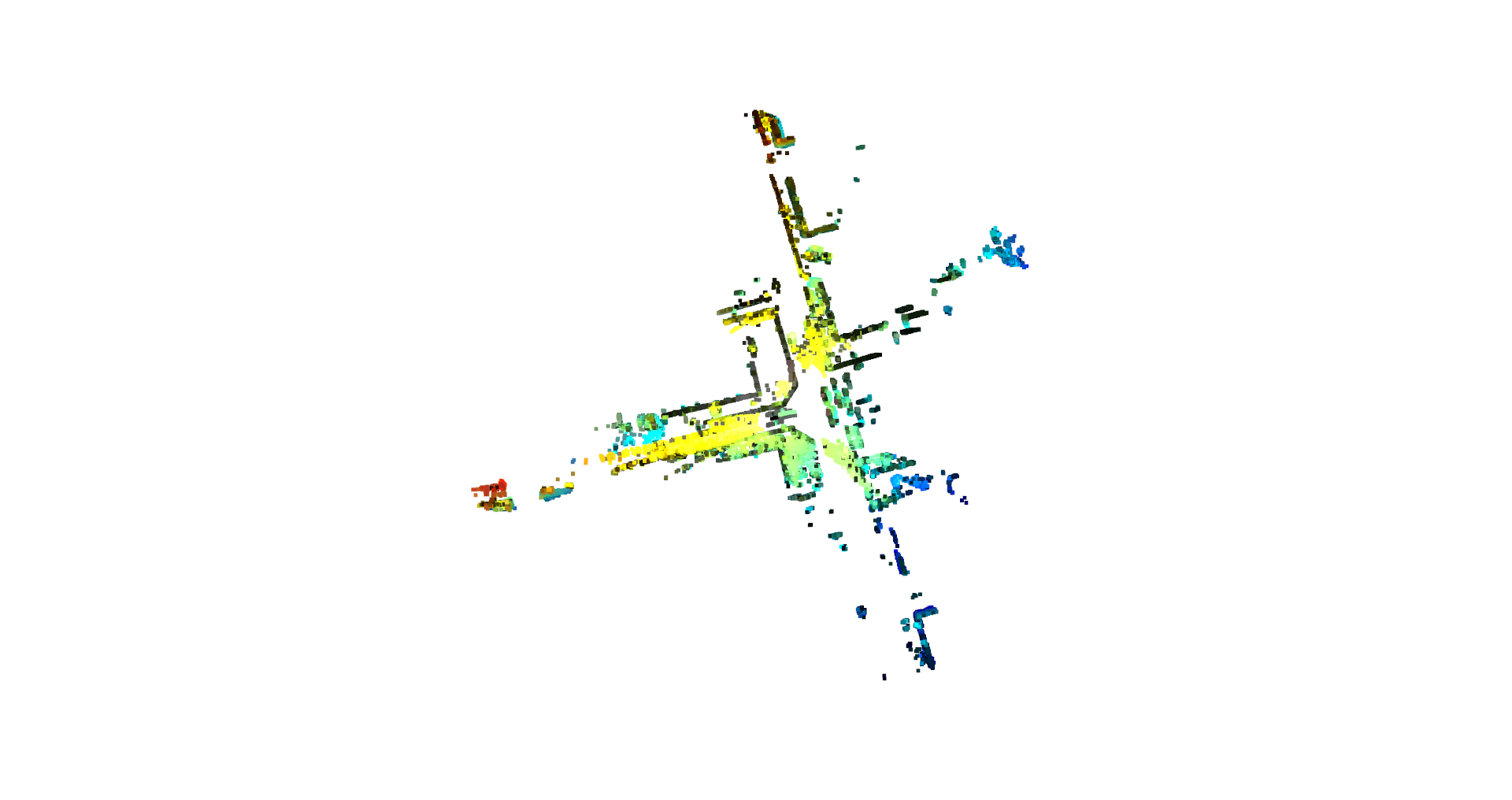}
    \vspace{0.1cm} 
    \includegraphics[width=0.99\linewidth]{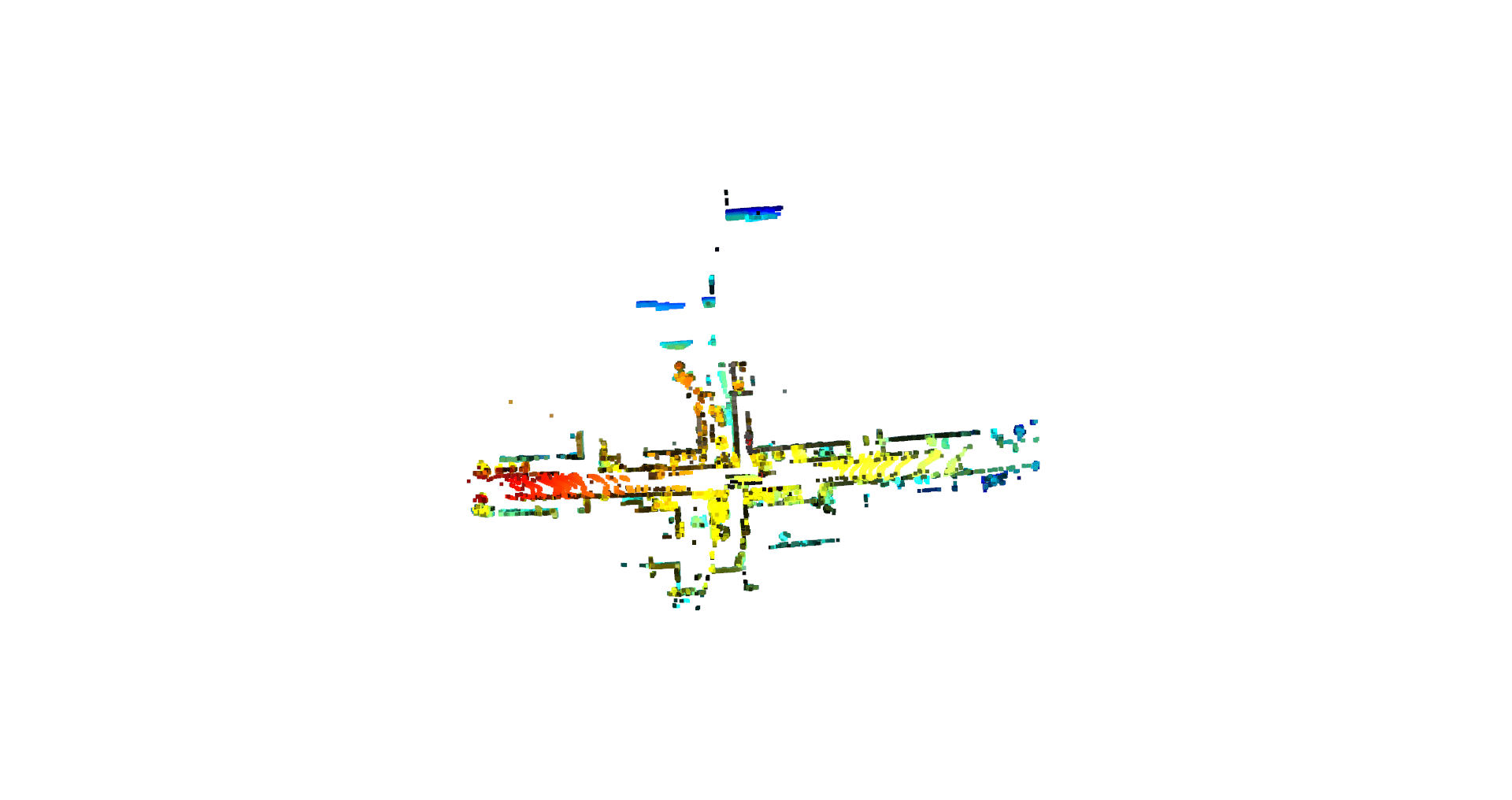}
    \vspace{0.1cm} 
    \includegraphics[width=0.99\linewidth]{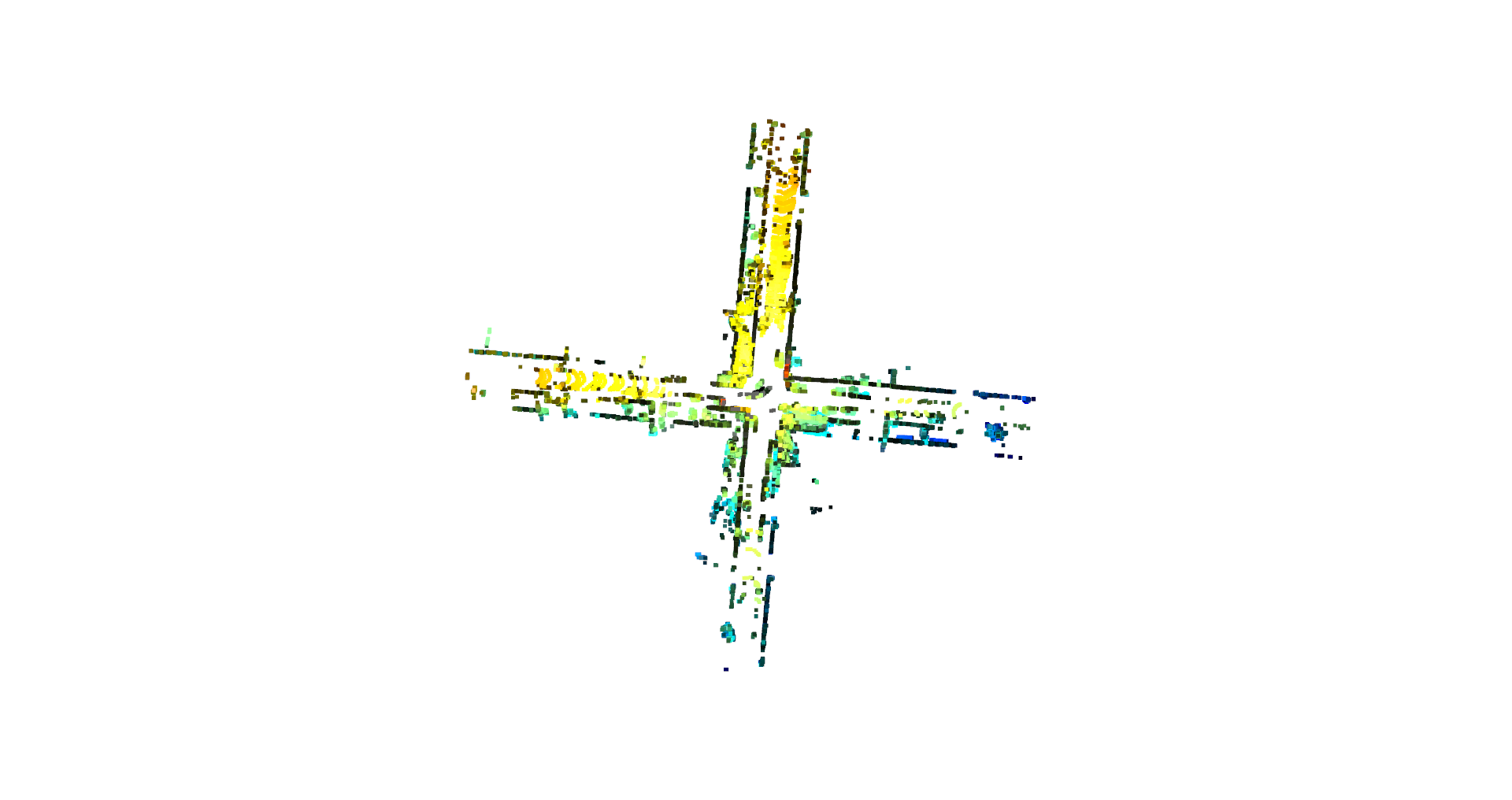}
    \vspace{0.1cm} 
    \includegraphics[width=0.99\linewidth]{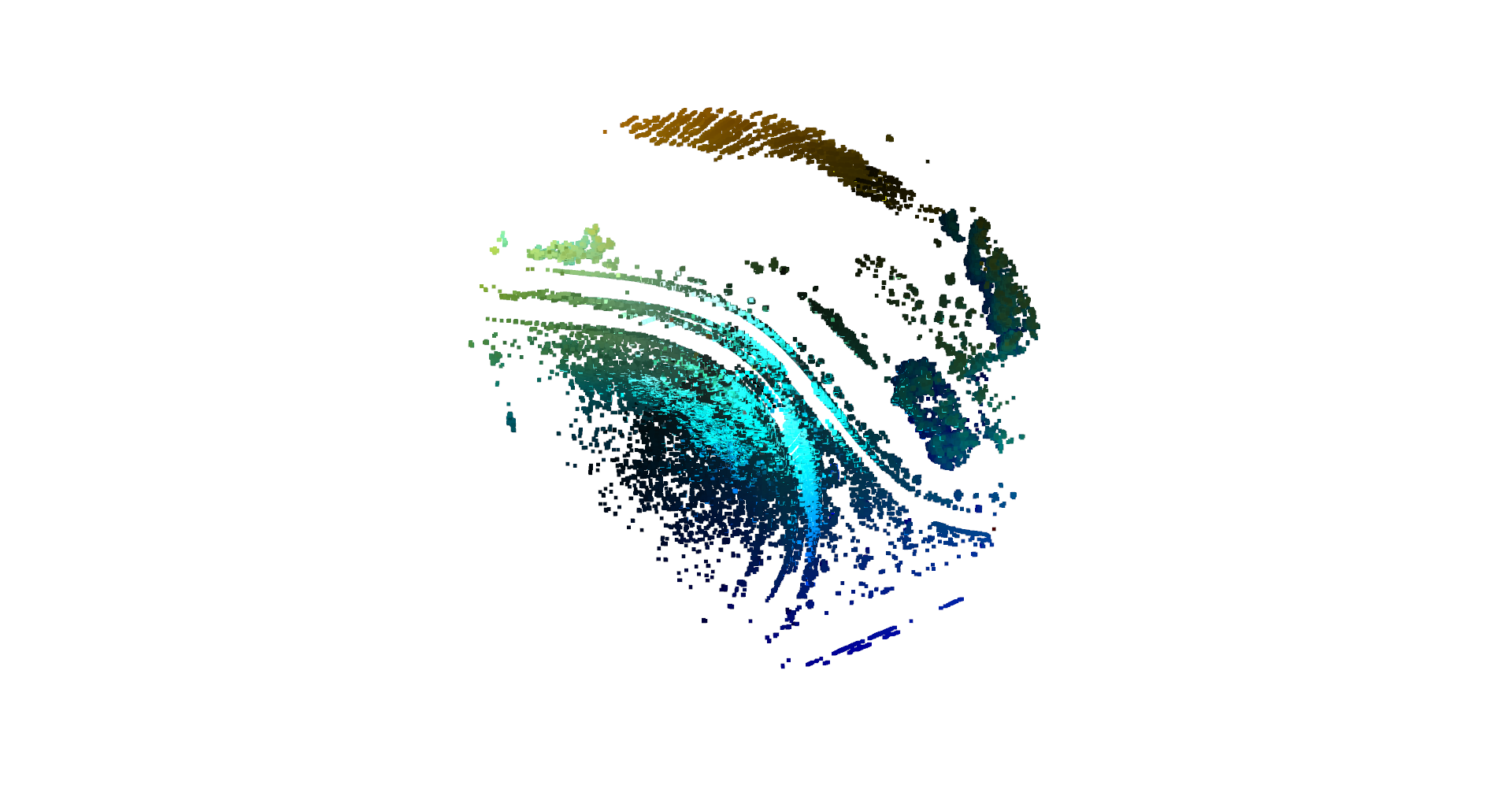}
  \end{minipage}
  \hfill
  \begin{minipage}[b]{0.24\textwidth}
    \centering
    \includegraphics[width=0.99\linewidth]{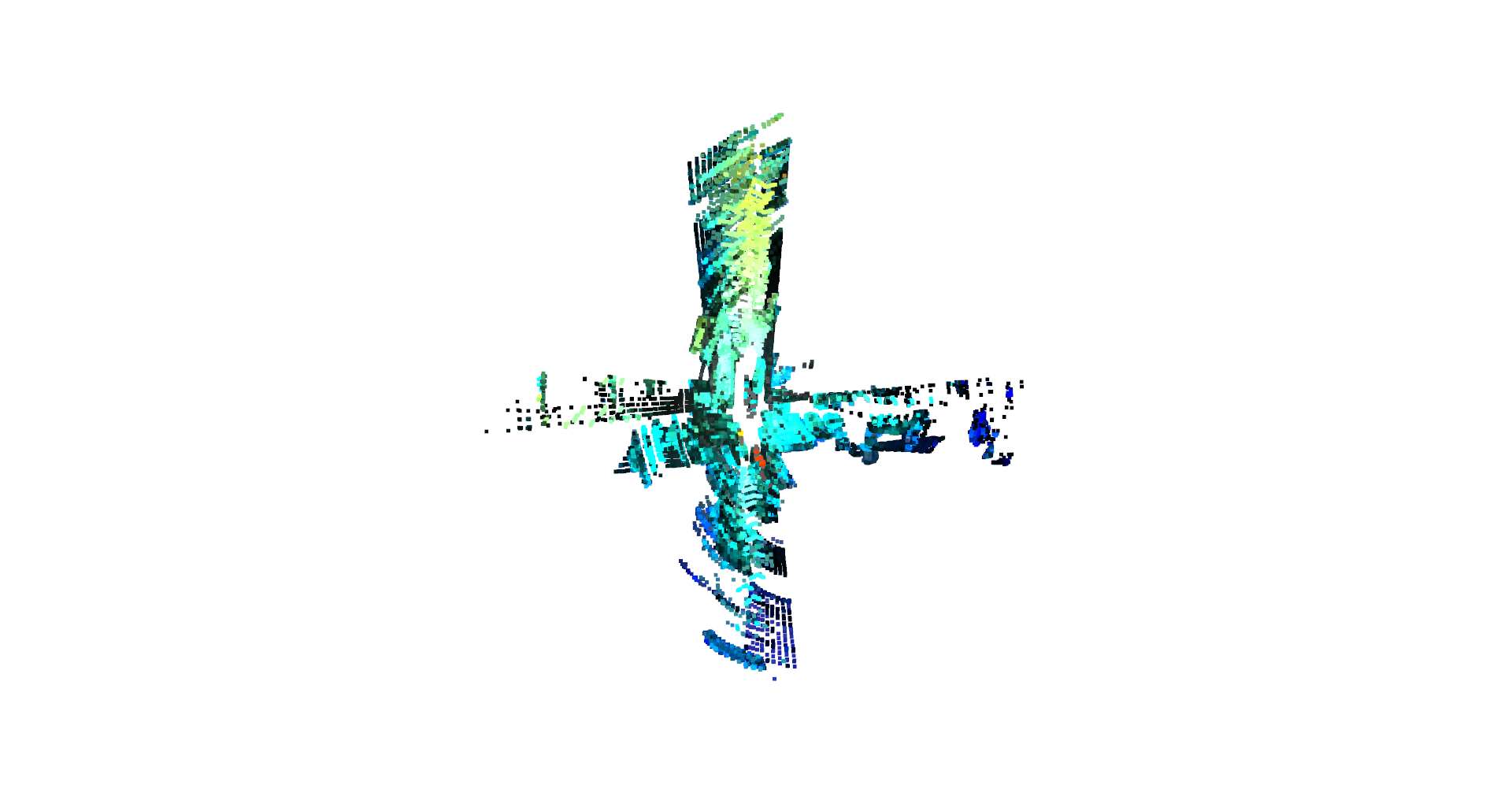}
    \vspace{0.1cm} 
    \includegraphics[width=0.99\linewidth]{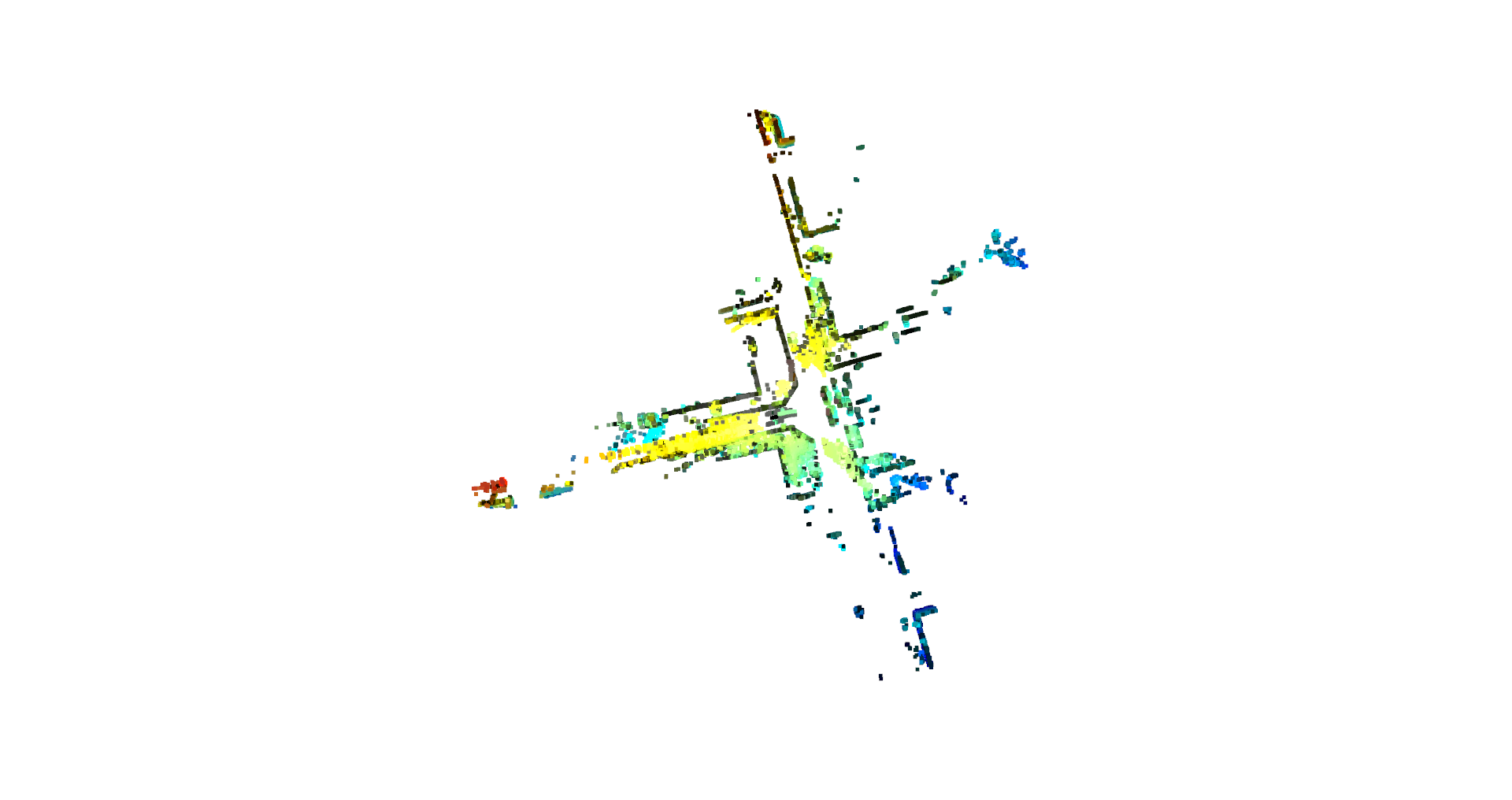}
    \vspace{0.1cm} 
    \includegraphics[width=0.99\linewidth]{figures/alligned_with_pred_1.png}
    \vspace{0.1cm} 
    \includegraphics[width=0.99\linewidth]{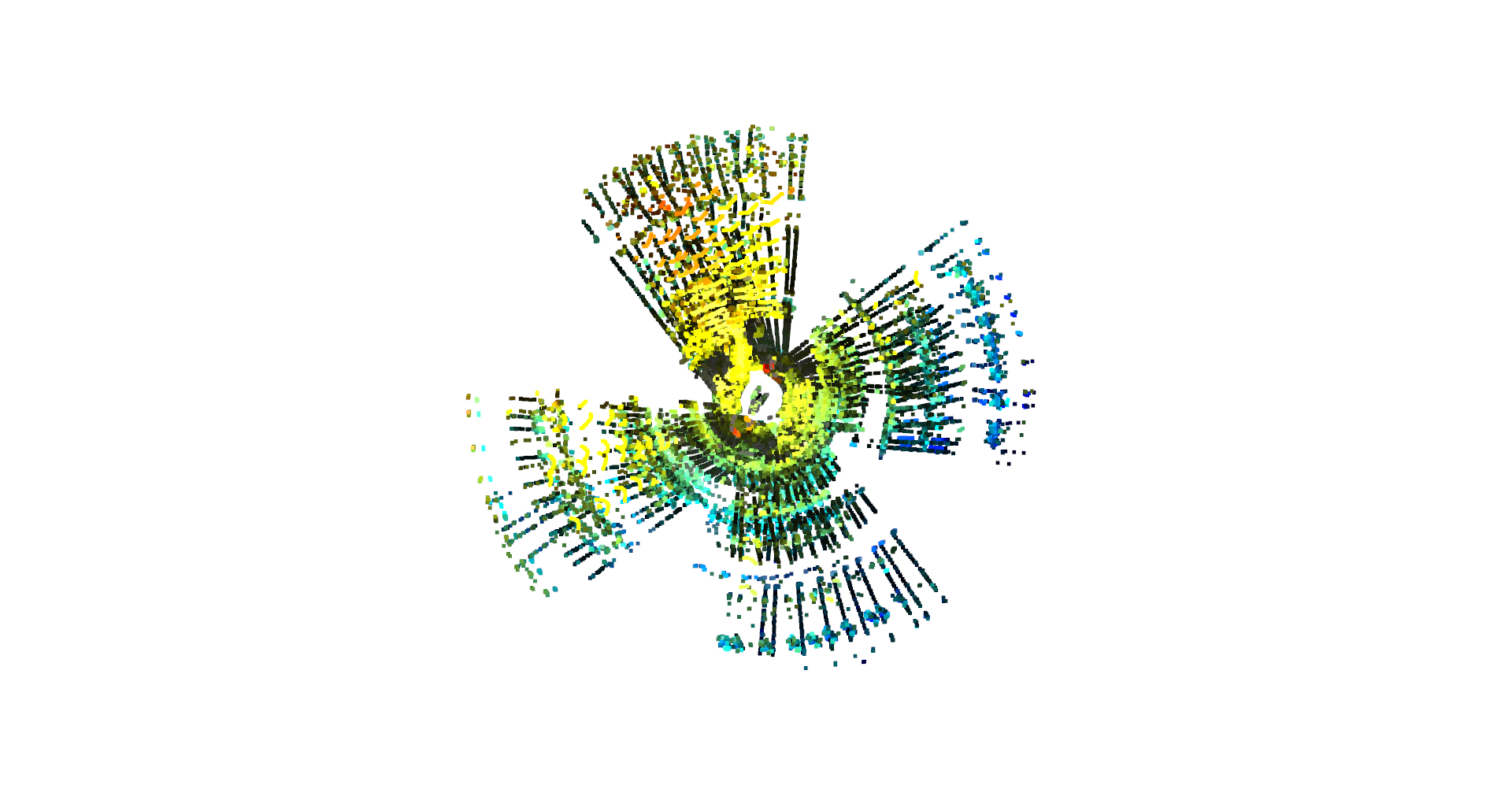}
        \vspace{0.1cm} 
    \includegraphics[width=0.99\linewidth]{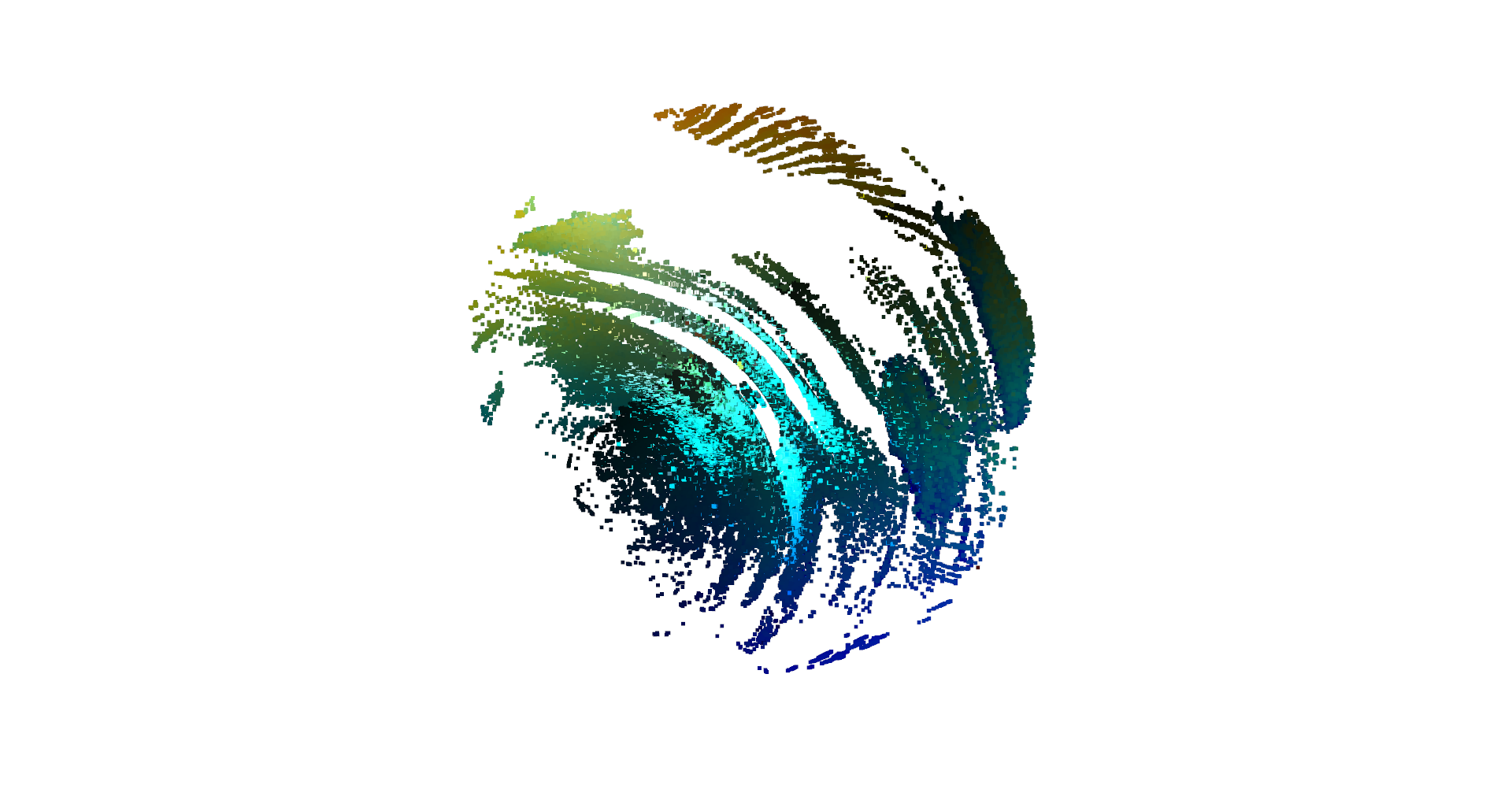}
  \end{minipage}
   \hfill
  \begin{minipage}[b]{0.24\textwidth}
    \centering
    \includegraphics[width=0.99\linewidth]{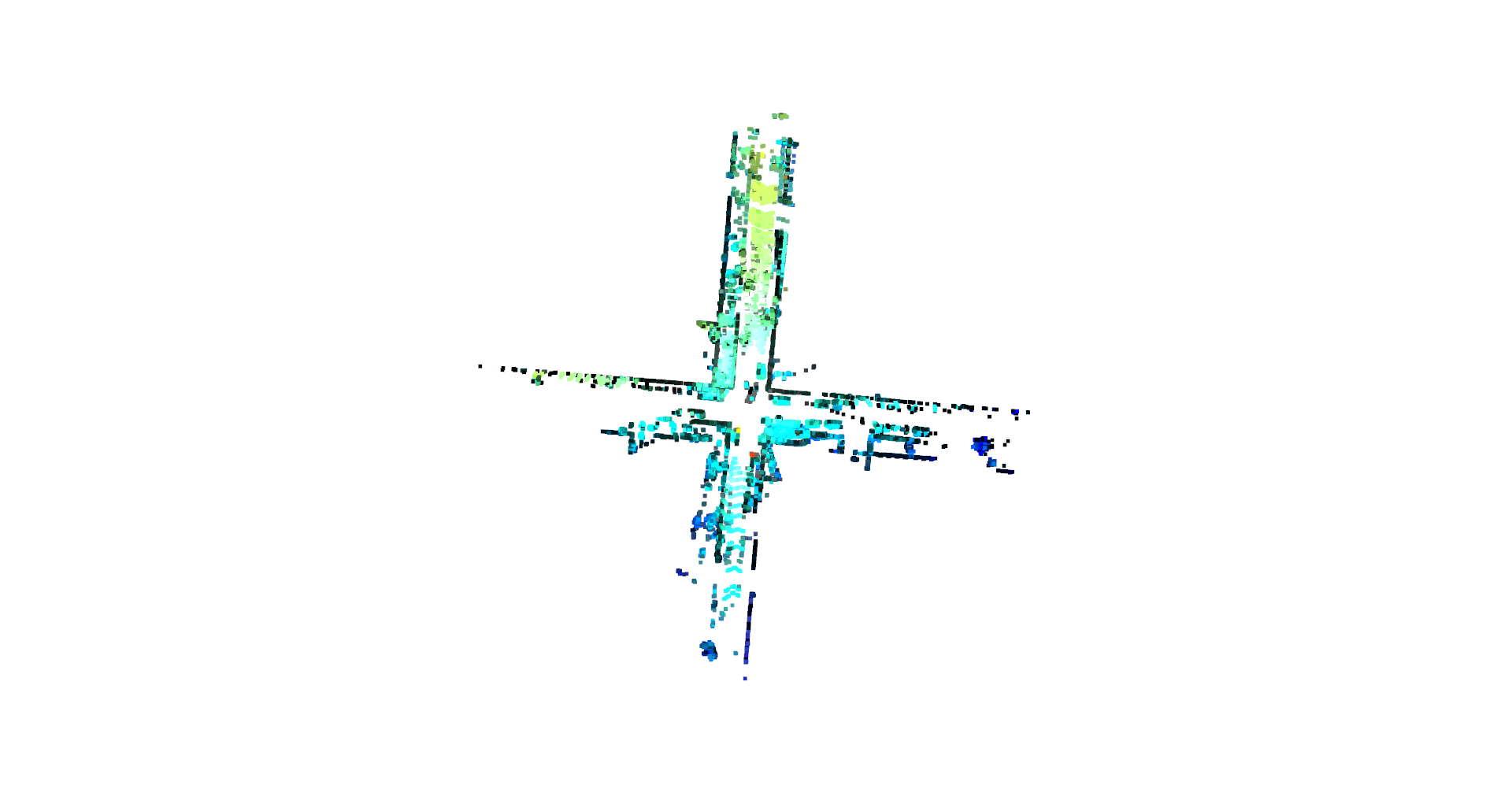}
    \vspace{0.1cm} 
    \includegraphics[width=0.99\linewidth]{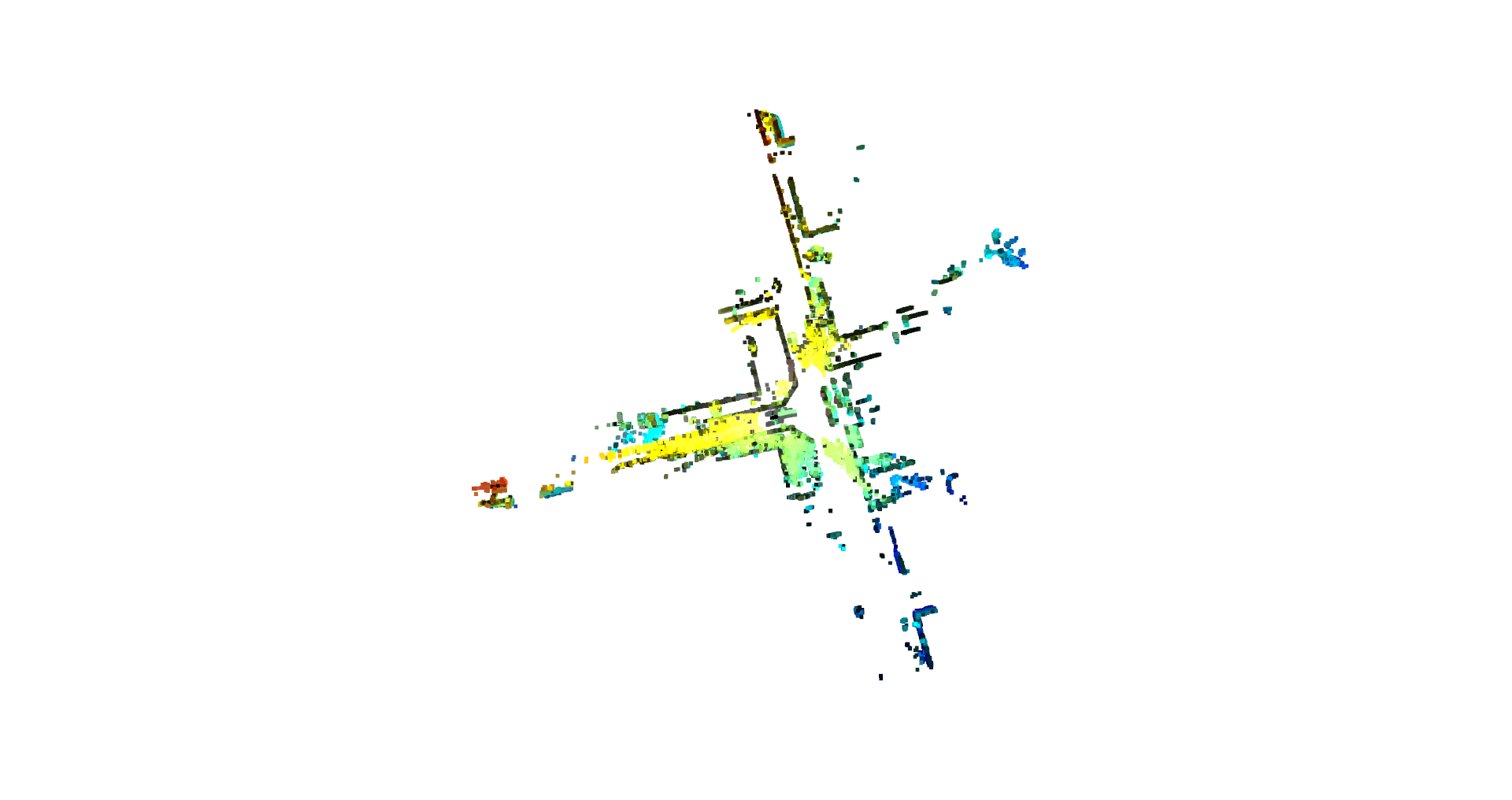}
    \vspace{0.1cm} 
    \includegraphics[width=0.99\linewidth]{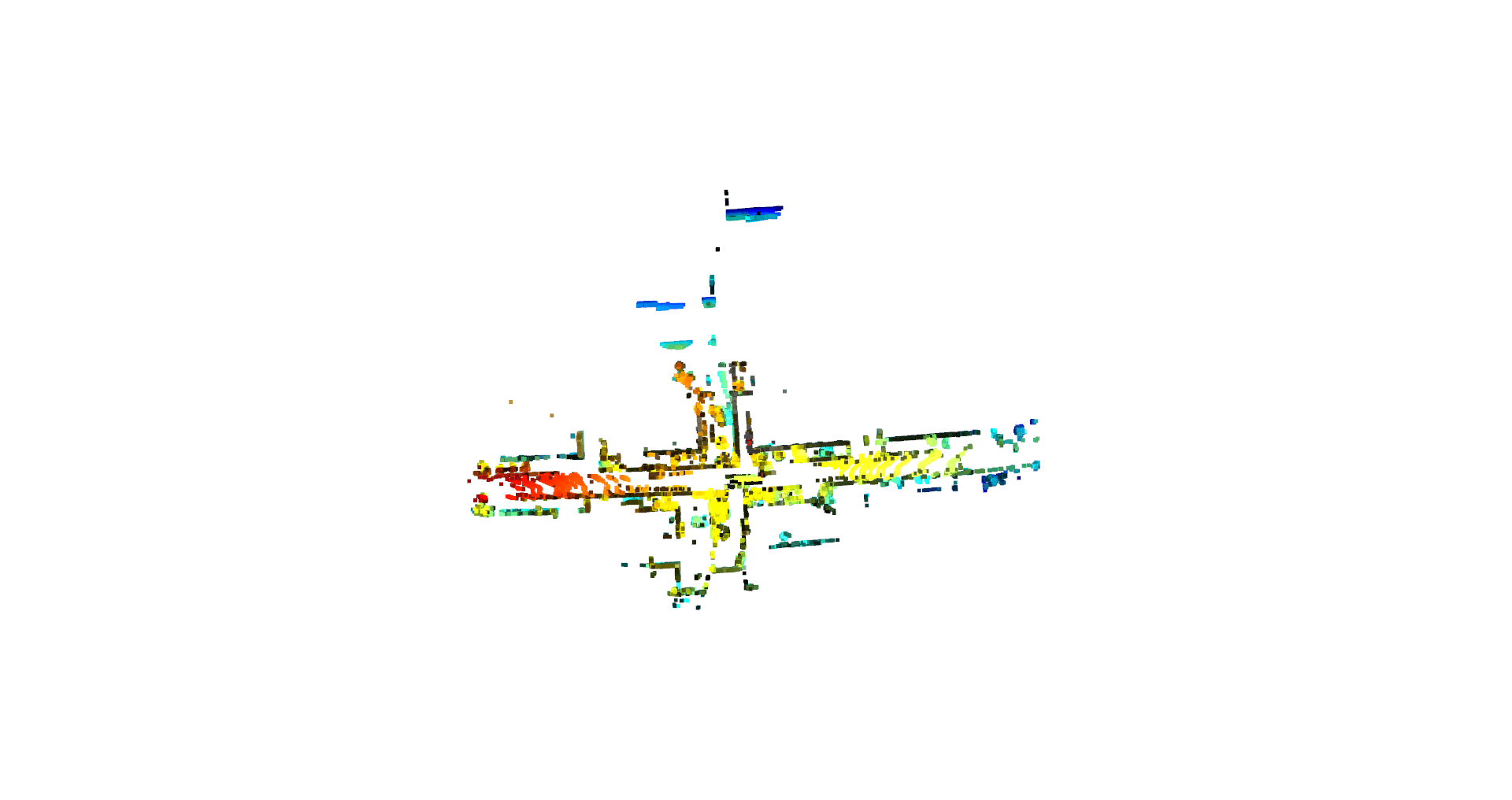}
    \vspace{0.1cm} 
    \includegraphics[width=0.99\linewidth]{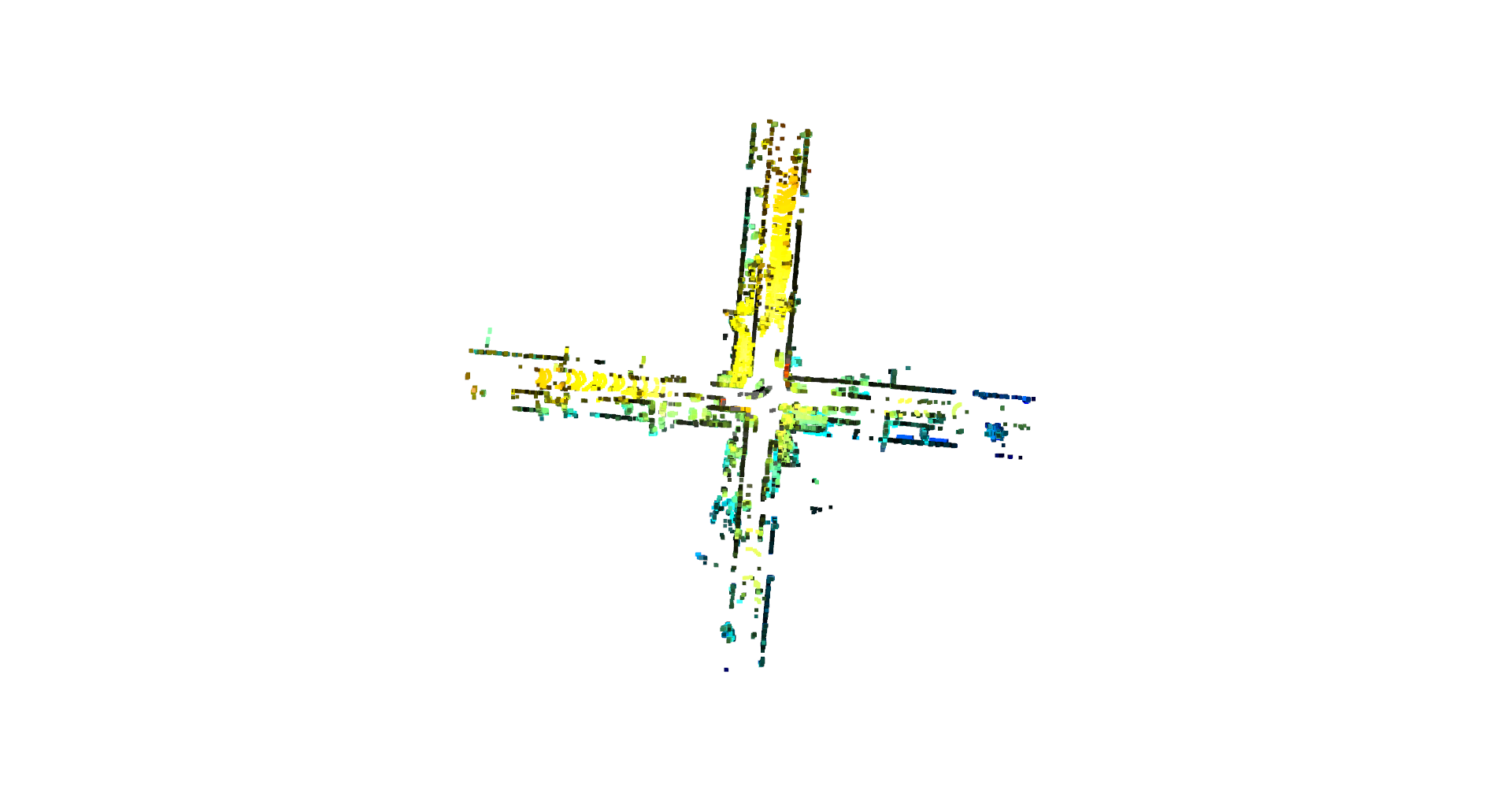}
    \vspace{0.1cm} 
    \includegraphics[width=0.99\linewidth]{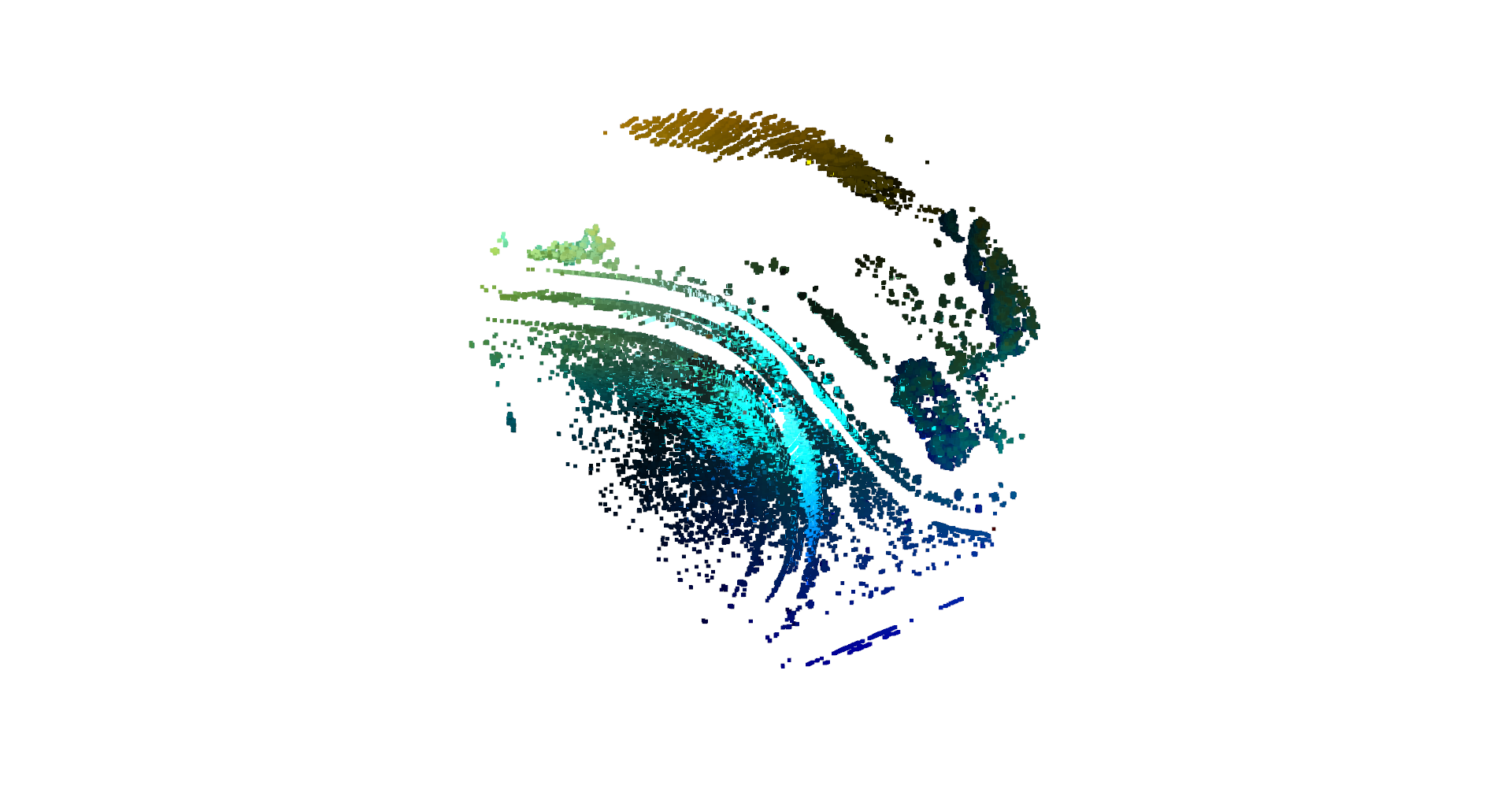}
  \end{minipage}
  
 \caption{From top to bottom, each row represents 10 point clouds derived from different sequences of the KITTI odometry dataset. From left to right: (1) Non-aligned scans; (2) Point clouds aligned using our estimated poses; (3) Point clouds aligned through multiway registration via pose-graph optimization; (4) Point clouds aligned according to ground truth poses.}

\label{fig:pcd-comparison}
  \label{fig:pcd-comparison}
\end{figure*}

In the majority of test sequences examined, both methods yielded a largely comparable translational error. However, our learned model presents significantly less rotational error.

Essentially, initializing the pose to estimate using the previous frame can indeed make the optimization process easier, especially when dealing with sequences of frames where changes between consecutive frames are relatively small. However, this method can encounter problems when there is significant rotational movement, where the assumption of pose similarity between successive frames may not hold, rendering the method less effective.

The trained model takes advantage of temporal dynamics, reducing its dependence on pose information from the previous data frame, a strategy that gives it better flexibility and adaptability in the search for a wide range of potential solutions. 


\section{Conclusion}
\label{sec:Discussion}

In this work, we presented a learned model for processing point clouds data and determining rigid poses through a deep learning architecture that integrates a dynamic graph neural network with a hierarchical attention mechanism. 
Our method demonstrated promising results in the rigid pose estimation task, effectively compressing high-dimensional data into a low-dimensional vector representation. This was achieved by leveraging hierarchical attention mechanism and maximum operation to extract the most crucial features of each point cloud. Moreover, we opted for a Gram-Schmidt orthonormalization parametrization in the pose decoding stage, which yielded more precise results.
This framework inherently supports a deep understanding of the spatial relationships and dynamics, leveraging both local and global contextual information in estimating ego-poses.


A significant insight drawn from our evaluations, conducted on the KITTI Odometry dataset, reveals that in many instances, our method outperforms the established multiway registration via pose graph optimization approach, notably reducing rotational error in several test sequences. This assert the potential effectiveness and applicability of our proposed model for geometric registration over large point clouds, particularly where rotational accuracy is of crucial importance.

In future work, we plan to further investigate the relevance of our model over larger windows and its direct usability within Simultaneous Localization and Mapping (SLAM) algorithm. This will involve applying the model to submaps or tiles \cite{melbouci2020lpg} that compile a larger number of scans, aggregated from odometry poses. Our objective is to replace the existing loop closure optimization scheme in SLAM with our learned model.

%
%
%
%

\bibliographystyle{IEEEtran}
\bibliography{IEEEabrv,bibfile}

\begin{thebibliography}{10}
\providecommand{\url}[1]{#1}
\csname url@rmstyle\endcsname
\providecommand{\newblock}{\relax}
\providecommand{\bibinfo}[2]{#2}
\providecommand\BIBentrySTDinterwordspacing{\spaceskip=0pt\relax}
\providecommand\BIBentryALTinterwordstretchfactor{4}
\providecommand\BIBentryALTinterwordspacing{\spaceskip=\fontdimen2\font plus
\BIBentryALTinterwordstretchfactor\fontdimen3\font minus
  \fontdimen4\font\relax}
\providecommand\BIBforeignlanguage[2]{{%
\expandafter\ifx\csname l@#1\endcsname\relax
\typeout{** WARNING: IEEEtran.bst: No hyphenation pattern has been}%
\typeout{** loaded for the language `#1'. Using the pattern for}%
\typeout{** the default language instead.}%
\else
\language=\csname l@#1\endcsname
\fi
#2}}

\bibitem{melbouci2020lpg}
K.~Melbouci and F.~Nashashibi, ``Lpg-slam: a light-weight probabilistic
  graph-based slam,'' in \emph{ICARCV 2020-International Conference on Control,
  Automation, Robotics and Vision}, 2020.

\bibitem{choi2015robust}
S.~Choi, Q.-Y. Zhou, and V.~Koltun, ``Robust reconstruction of indoor scenes,''
  in \emph{Proceedings of the IEEE conference on computer vision and pattern
  recognition}, 2015, pp. 5556--5565.

\bibitem{vaswani2017attention}
A.~Vaswani, N.~Shazeer, N.~Parmar, J.~Uszkoreit, L.~Jones, A.~N. Gomez,
  {\L}.~Kaiser, and I.~Polosukhin, ``Attention is all you need,''
  \emph{Advances in neural information processing systems}, vol.~30, 2017.

\bibitem{https://doi.org/10.48550/arxiv.2207.09238}
\BIBentryALTinterwordspacing
M.~Phuong and M.~Hutter, ``Formal algorithms for transformers,'' 2022.
  [Online]. Available: \url{https://arxiv.org/abs/2207.09238}
\BIBentrySTDinterwordspacing

\bibitem{guo2022attention}
M.-H. Guo, T.-X. Xu, J.-J. Liu, Z.-N. Liu, P.-T. Jiang, T.-J. Mu, S.-H. Zhang,
  R.~R. Martin, M.-M. Cheng, and S.-M. Hu, ``Attention mechanisms in computer
  vision: A survey,'' \emph{Computational Visual Media}, pp. 1--38, 2022.

\bibitem{khan2022transformers}
S.~Khan, M.~Naseer, M.~Hayat, S.~W. Zamir, F.~S. Khan, and M.~Shah,
  ``Transformers in vision: A survey,'' \emph{ACM computing surveys (CSUR)},
  vol.~54, no. 10s, pp. 1--41, 2022.

\bibitem{ryoo2021tokenlearner}
M.~S. Ryoo, A.~Piergiovanni, A.~Arnab, M.~Dehghani, and A.~Angelova,
  ``Tokenlearner: What can 8 learned tokens do for images and videos?''
  \emph{arXiv preprint arXiv:2106.11297}, 2021.

\bibitem{924423}
S.~Rusinkiewicz and M.~Levoy, ``Efficient variants of the icp algorithm,'' in
  \emph{Proceedings Third International Conference on 3-D Digital Imaging and
  Modeling}, 2001, pp. 145--152.

\bibitem{5152473}
R.~B. Rusu, N.~Blodow, and M.~Beetz, ``Fast point feature histograms (fpfh) for
  3d registration,'' in \emph{2009 IEEE International Conference on Robotics
  and Automation}, 2009, pp. 3212--3217.

\bibitem{zhou2016fast}
Q.-Y. Zhou, J.~Park, and V.~Koltun, ``Fast global registration,'' \emph{arXiv
  preprint arXiv:1607.03557}, 2016.

\bibitem{kendall2015posenet}
A.~Kendall, M.~Grimes, and R.~Cipolla, ``Posenet: A convolutional network for
  real-time 6-dof camera relocalization,'' in \emph{Proceedings of the IEEE
  international conference on computer vision}, 2015, pp. 2938--2946.

\bibitem{guo2014deep}
X.~Guo, S.~Singh, H.~Lee, R.~L. Lewis, and X.~Wang, ``Deep learning for
  real-time atari game play using offline monte-carlo tree search planning,''
  \emph{Advances in neural information processing systems}, vol.~27, 2014.

\bibitem{huang2021comprehensive}
X.~Huang, G.~Mei, J.~Zhang, and R.~Abbas, ``A comprehensive survey on point
  cloud registration,'' \emph{arXiv preprint arXiv:2103.02690}, 2021.

\bibitem{qi2017pointnet}
C.~R. Qi, H.~Su, K.~Mo, and L.~J. Guibas, ``Pointnet: Deep learning on point
  sets for 3d classification and segmentation,'' in \emph{Proceedings of the
  IEEE conference on computer vision and pattern recognition}, 2017, pp.
  652--660.

\bibitem{park2022vision}
N.~Park and S.~Kim, ``How do vision transformers work?'' 2022.

\bibitem{wang2019deep}
Y.~Wang and J.~M. Solomon, ``Deep closest point: Learning representations for
  point cloud registration,'' in \emph{Proceedings of the IEEE/CVF
  international conference on computer vision}, 2019, pp. 3523--3532.

\bibitem{yu2021cofinet}
H.~Yu, F.~Li, M.~Saleh, B.~Busam, and S.~Ilic, ``Cofinet: Reliable
  coarse-to-fine correspondences for robust pointcloud registration,''
  \emph{Advances in Neural Information Processing Systems}, vol.~34, pp.
  23\,872--23\,884, 2021.

\bibitem{zhang20233d}
X.~Zhang, J.~Yang, S.~Zhang, and Y.~Zhang, ``3d registration with maximal
  cliques,'' in \emph{Proceedings of the IEEE/CVF Conference on Computer Vision
  and Pattern Recognition}, 2023, pp. 17\,745--17\,754.

\bibitem{hu2022nrtnet}
X.~Hu, D.~Zhang, J.~Chen, Y.~Wu, and Y.~Chen, ``Nrtnet: An unsupervised method
  for 3d non-rigid point cloud registration based on transformer,''
  \emph{Sensors}, vol.~22, no.~14, p. 5128, 2022.

\bibitem{slimani2023rocnet}
K.~Slimani, B.~Tamadazte, and C.~Achard, ``Rocnet: 3d robust registration of
  point-clouds using deep learning,'' \emph{arXiv preprint arXiv:2303.07963},
  2023.

\bibitem{arce2023padloc}
J.~Arce, N.~V{\"o}disch, D.~Cattaneo, W.~Burgard, and A.~Valada, ``Padloc:
  Lidar-based deep loop closure detection and registration using panoptic
  attention,'' \emph{IEEE Robotics and Automation Letters}, vol.~8, no.~3, pp.
  1319--1326, 2023.

\bibitem{shi2020pv}
S.~Shi, C.~Guo, L.~Jiang, Z.~Wang, J.~Shi, X.~Wang, and H.~Li, ``Pv-rcnn:
  Point-voxel feature set abstraction for 3d object detection,'' in
  \emph{Proceedings of the IEEE/CVF conference on computer vision and pattern
  recognition}, 2020, pp. 10\,529--10\,538.

\bibitem{li2022lepard}
Y.~Li and T.~Harada, ``Lepard: Learning partial point cloud matching in rigid
  and deformable scenes,'' in \emph{Proceedings of the IEEE/CVF conference on
  computer vision and pattern recognition}, 2022, pp. 5554--5564.

\bibitem{thomas2019kpconv}
H.~Thomas, C.~R. Qi, J.-E. Deschaud, B.~Marcotegui, F.~Goulette, and L.~J.
  Guibas, ``Kpconv: Flexible and deformable convolution for point clouds,'' in
  \emph{Proceedings of the IEEE/CVF international conference on computer
  vision}, 2019, pp. 6411--6420.

\bibitem{qin2022geometric}
Z.~Qin, H.~Yu, C.~Wang, Y.~Guo, Y.~Peng, and K.~Xu, ``Geometric transformer for
  fast and robust point cloud registration,'' in \emph{Proceedings of the
  IEEE/CVF conference on computer vision and pattern recognition}, 2022, pp.
  11\,143--11\,152.

\bibitem{monji2023review}
S.~Monji-Azad, J.~Hesser, and N.~L{\"o}w, ``A review of non-rigid
  transformations and learning-based 3d point cloud registration methods,''
  \emph{ISPRS Journal of Photogrammetry and Remote Sensing}, vol. 196, pp.
  58--72, 2023.

\bibitem{trappolini2021shape}
G.~Trappolini, L.~Cosmo, L.~Moschella, R.~Marin, S.~Melzi, and E.~Rodol{\`a},
  ``Shape registration in the time of transformers,'' \emph{Advances in Neural
  Information Processing Systems}, vol.~34, pp. 5731--5744, 2021.

\bibitem{wang2019dynamic}
Y.~Wang, Y.~Sun, Z.~Liu, S.~E. Sarma, M.~M. Bronstein, and J.~M. Solomon,
  ``Dynamic graph cnn for learning on point clouds,'' \emph{Acm Transactions On
  Graphics (tog)}, vol.~38, no.~5, pp. 1--12, 2019.

\bibitem{shen2021accurate}
Z.~Shen, J.~Feydy, P.~Liu, A.~H. Curiale, R.~San Jose~Estepar, R.~San
  Jose~Estepar, and M.~Niethammer, ``Accurate point cloud registration with
  robust optimal transport,'' \emph{Advances in Neural Information Processing
  Systems}, vol.~34, pp. 5373--5389, 2021.

\bibitem{guo2021pct}
M.-H. Guo, J.-X. Cai, Z.-N. Liu, T.-J. Mu, R.~R. Martin, and S.-M. Hu, ``Pct:
  Point cloud transformer,'' \emph{Computational Visual Media}, vol.~7, no.~2,
  pp. 187--199, 2021.

\bibitem{zhou2019continuity}
Y.~Zhou, C.~Barnes, J.~Lu, J.~Yang, and H.~Li, ``On the continuity of rotation
  representations in neural networks,'' in \emph{Proceedings of the IEEE/CVF
  Conference on Computer Vision and Pattern Recognition}, 2019, pp. 5745--5753.

\bibitem{bregier2021deep}
R.~Br{\'e}gier, ``Deep regression on manifolds: a 3d rotation case study,'' in
  \emph{2021 International Conference on 3D Vision (3DV)}.\hskip 1em plus 0.5em
  minus 0.4em\relax IEEE, 2021, pp. 166--174.

\bibitem{Zhou2018}
Q.-Y. Zhou, J.~Park, and V.~Koltun, ``{Open3D}: {A} modern library for {3D}
  data processing,'' \emph{arXiv:1801.09847}, 2018.

\bibitem{Geiger2012CVPR}
A.~Geiger, P.~Lenz, and R.~Urtasun, ``Are we ready for autonomous driving? the
  kitti vision benchmark suite,'' in \emph{Conference on Computer Vision and
  Pattern Recognition (CVPR)}, 2012.

\end{thebibliography}
\end{document}